\documentclass[]{interact}

\usepackage{epstopdf}
\usepackage{subfigure}

\usepackage{float}
\usepackage{amsmath}
\restylefloat{figure}
\usepackage{multirow}
\usepackage{graphics}
\usepackage{hyperref}
\usepackage{cleveref}
\usepackage{rotating}
\usepackage{soul}
\usepackage{longtable}
\usepackage{ltablex}
\usepackage{caption, booktabs}

\usepackage{natbib}
\bibpunct[, ]{(}{)}{;}{a}{}{,}
\renewcommand\bibfont{\fontsize{10}{12}\selectfont}

\theoremstyle{plain}

\theoremstyle{definition}

\theoremstyle{remark}

\begin{document}


\title{A mathematical model for simultaneous personnel shift planning and unrelated parallel machine scheduling}

\author{
\name{Maziyar Khadivi\textsuperscript{a}, Mostafa Abbasi\textsuperscript{a}, Todd Charter\textsuperscript{b} and Homayoun Najjaran\textsuperscript{a,b} \thanks{CONTACT H. Najjaran Email: najjaran@uvic.ca}}
\affil{\textsuperscript{a}Department of Mechanical Engineering, University of Victoria, Victoria BC, V8P 5C2, Canada \textsuperscript{b}Department of Electrical and Computer Engineering, University of Victoria, Victoria BC, V8P 5C2, Canada}
}

\maketitle

\begin{abstract}
This paper addresses a production scheduling problem derived from an industrial use case, focusing on unrelated parallel machine scheduling with the personnel availability constraint. The proposed model optimizes the production plan over a multi-period scheduling horizon, accommodating variations in personnel shift hours within each time period. It assumes shared personnel among machines, with one personnel required per machine for setup and supervision during job processing. Available personnel are fewer than the machines, thus limiting the number of machines that can operate in parallel. The model aims to minimize the total production time considering machine-dependent processing times and sequence-dependent setup times. The model handles practical scenarios like machine eligibility constraints and production time windows. A Mixed Integer Linear Programming (MILP) model is introduced to formulate the problem, taking into account both continuous and district variables. A two-step solution approach enhances computational speed, first maximizing accepted jobs and then minimizing production time. Validation with synthetic problem instances and a real industrial case study of a food processing plant demonstrates the performance of the model and its usefulness in personnel shift planning. The findings offer valuable insights for practical managerial decision-making in the context of production scheduling.
\end{abstract}


\section{Introduction}

 \label{sec: introduction}

Various types of machine scheduling problems exist, each tailored to model how jobs are processed by machines within a specific environment \citep{shojaeinasab2022intelligent}. Unrelated Parallel Machine (UPM) is one of the most studied machine scheduling problems in the literature. In general, UPM can be used to model different scheduling problems common in practice, including production scheduling in manufacturing, hospital operating room scheduling, call center workforce scheduling, airport runway scheduling, and parallel job scheduling in supercomputing \citep{burdett2021scheduling}. In this study, we explore the application of UPM to solve production scheduling in manufacturing systems. In the manufacturing domain, UPM is suitable to model high output production shops or the most critical stage in certain production processes, for instance, the assembly stage in car manufacturing industry, packing hall operations in food processing facilities, or the kiln firing stage in ceramic tile manufacturing \citep{fanjul2019reformulations, ahang2024intelligent}. It involves the allocation and sequencing of $n$ independent jobs across $m$ machines that perform the same task in parallel, but possibly at different speeds \citep{khadivi2023deep}. The optimization of these allocation and sequencing decisions is inherently tied to the Key Performance Indicators (KPIs) pursued by the manufacturing enterprise or policies \citep{abbasi2020land}, including production makespan, production volume, and order fulfillment \citep{liang2022lenovo}. In many real-world settings, machines may require cleaning, reconfigurations, color preparations, or adjustments prior to processing the next job, referred to as changeover times or setup times. Setup times are nonproductive activities occurring between the production of two consecutive jobs on the same machine. We consider the most general format of setup times in which the setup duration depends on both the machine and the sequence of jobs \citep{fanjul2019reformulations}. In the literature, this problem is referred to as Unrelated Parallel Machine scheduling problem with Setup times (UPMS).\\

In many real-world production scheduling problems, auxiliary resources are often required for the setup of machines and processing of jobs. These resources might be available in limited quantities and include machine operators, dies, tools, pallets, and industrial robots \citep{afzalirad2016resource, ogunfowora2023reinforcement}. Overlooking the auxiliary resource constraint can lead to generating solutions that are not feasible to implement in reality. In the literature, authors referred to this category of problem either as resource constrained UPMS or as UPMS with additional$/$auxiliary resources (UPMSR) \citep{fanjul2020models}. According to \cite{blazewicz2007handbook}, an additional resource can only be only needed during the processing of a job on the machines (processing resource) or prior and after the processing of a job during machine setups, called input–output resource. We consider the cases in which the resource is required for both job processing and machine setups. Furthermore, consistent with the classification provided by \cite{edis2013parallel} for different types of resources, we focus on a dynamic doubly constrained discrete single resource. Being dynamic implies that the allocation of resources is not fixed prior to job scheduling and it is rather determined while solving the machine scheduling problem. Although this would significantly increase the complexity of the problem, it results in considerable improvements in shortening the production makespan as a resource can dynamically move among machines, as reported by \cite{daniels1996scheduling,daniels1997analysis}. Doubly constrained indicates that the resource is both renewable and nonrenewable. A resource is considered renewable when its total amount is limited at any given time and after being used for a task, it can be assigned to another. At the same time, a resource can be nonrenewable when its total usage is constrained and after being consumed by a task, its availability will be reduced \Citep{slowinski1980two, blazewicz2007handbook}. In terms of divisibility, a resource can be assigned to a task in discrete values from a finite set of available resources or in continuous values from flexible quantities within a given interval. Our study focuses on discrete resources. Lastly, we consider a single type of additional resource, which is manpower in our use case, for both job processing and machine setup. The number of resources required per machine is fixed to one and identical across all machines. An example of a doubly constrained discrete resource is the use of human operators that can be required in manufacturing plants as an auxiliary resource to set up and run machines. The human operator is a discrete resource and can be limited in quantity, i.e., a renewable resource. Moreover, their availability in the given time periods (e,g., weeks) can be treated as a continuous value but possibly with a varying amount in each time period, i.e., a nonrenewable resource.\\


In the literature, previous works have studied parallel machine scheduling problems with additional resources required for job processing \citep{afzalirad2016resource,al2019optimize, yunusoglu2022constraint, chen2022unrelated, bitar2021unrelated}, machine setups \citep{costa2013hybrid, yepes2020grasp}, or both \citep{fanjul2020models}. \textbf{Different from previous studies}, our work contributes to the literature by accounting for the variability in the availability of the resources, which can often be the case in reality. This contribution aligns with the future research direction identified by \cite{fanjul2019reformulations}, who emphasized the need to consider additional manufacturing resources, such as personnel whose availability can be as limited as machines; therefore, requiring proper assignment. Our mathematical model formulation takes into account the most general format of parallel machines, UPMS. The model also involves real-world manufacturing constraints related to machine eligibility, job release times, and job delivery times. The proposed model is based on Mixed Integer Linear Programming (MILP). The MILP model uses a time-continuous formulation to deal with the scheduling of jobs and auxiliary resources on the machines. As shown by previous work, e.g., \citep{castro2006two,afzalirad2016resource, fanjul2019reformulations, tahmassebi1999vehicle, avalos2015efficient}, this formulation yields solutions with less optimality gap, faster solving speed, and higher scalability compared to others, such as the time-indexed or discrete time formulation. The model also proposes using the novel idea of discrete "Positions" for allocation of personnel to set up and supervise the machine during processing the jobs. The concept of position together with the time-continuous formulation enables involving job release/delivery times and variable availability of each personnel in the planning horizon. To speed up execution time, we propose a Two-Step Solution Approach (TSSA) to solve the problem. The first step aims to find an initial feasible solution as a warm-start solution for the second step. The first step attempts to accommodate production of all given jobs through maximizing their total allocation, while the second step aims to minimize the total production time. The results of computational experiments indicate that the model formulation along with the two-step solution approach can solve problem instances with 120 jobs, 8 machines, and 7 personnel.\\

The remaining sections of the paper are organized as follows: \Cref{sec: literature} provides a review of the literature relevant to the scheduling problem under investigation. \Cref{sec: Problem definition} outlines the problem definition and provides examples for better understanding the basic idea behind the mathematical model. \Cref{sec: MIP} introduces the mathematical model for the problem along with the two-step solution approach. The procedure for generating synthetic problem instances and the information about the case study are provided in \Cref{sec: computational experiments}. The results of the computational experiments on the synthetically generated problem instances together with the case study are discussed in \Cref{sec: Results}. Finally, \Cref{sec: conclusions} outlines conclusions and potential avenues for future research.

\section{Literature review} \label{sec: literature}

In the literature pertinent to UPMS with additional resources, a group of studies assumed that auxiliary resources are necessary only when jobs are processed on the machines. \cite{afzalirad2016resource} approached the block erection scheduling problem in a shipyard as UPMS. They assumed a set of resources, such as manpower, industrial robots, tools, and dies, was required during job processing. They also considered manufacturing constraints, including job release dates, job precedence, and machine eligibility. The authors developed a nonlinear MIP model with time-indexed variables and two metaheuristics to solve the problem. \cite{al2019optimize} tackled the long solution time of the nonlinear model proposed by \cite{afzalirad2016resource} by introducing a linear time-indexed MILP model. They also presented a two-stage hybrid metaheuristic that combined variable neighborhood search with simulated annealing to solve the given problem with 60 jobs, 8 machines, and 8 resources. \cite{yunusoglu2022constraint} proposed a Constraint Programming (CP) model with different branching strategies to obtain optimal solutions for the problem defined by \cite{afzalirad2016resource}. The CP model could obtain optimal schedules for problem instances with 60 jobs, 8 machines, and 2 resources. \cite{bitar2021unrelated} developed a MILP model with time-indexed variables to optimize UPMS with auxiliary resources, required during job processing. With the aim of minimizing the completion time, this model optimized problem instances with up to 25 jobs, 3 machines, and 12 resources within 15 minutes. \cite{chen2022unrelated} modelled the photolithography stage of wafer production in semiconductor manufacturing as UPMS, where reticles were essential as an auxiliary resource for processing jobs on the machines. They presented two time-continuous MILP models, one based on network variables and another based on allocation and sequencing variables, to solve the problem considering job release times, sequence-dependent setup-time of resources (reticles) and location-dependent transfer-time of the resources. Their results show that modeling based on allocation and sequencing variables performs better when evaluated on problem sizes with 20 jobs, 3 machines, and 5 resources. To efficiently solve problems with 100 jobs, 15 machines, and 90 resources within a short time frame (10 minutes), they also designed a metaheuristic algorithm, called naked mole-rat.\\

\begin{table}[ht]
\caption{Summary of previous works employing exact methods for resource-constrained UPMS}
\label{tab:summaries}
\resizebox{\columnwidth}{!}{
\centering
\begin{tabular}{m{3.5cm}m{2.5cm}m{2cm}m{2.5cm}m{2.5cm}m{2.25cm}m{3cm}m{1.5cm}m{2.5cm}} 

\toprule

Authors (Year) & Resource type & Resource availability & Resource need per each job/machine & Number of resources & Scheduling horizon & Considerations & Exact method & Max. problem size $(Job \times Machine \times Resource)$ \\

\midrule

\citet{costa2013hybrid} & Machine setup & Constant & Fixed (one per setup) & Single resource (personnel) & Single period & - & MINLP & $(10 \times 5 \times 3)$\\

\citet{afzalirad2016resource} & Job processing & Constant & Variable & Multiple resources & Single period & Release date, machine eligibility, and job precedence constraint & MINLP & $(6 \times 3 \times 2)$\\

\citet{al2019optimize} & Job processing & Constant & Variable & Multiple resources & Single period & Release date & MILP & $(6 \times 2 \times 2)$\\

\citet{bektur2019mathematical} & Machine setup & Constant & Fixed (one per machine) & Single resource (Worker) & Single period & - & MILP & $(25 \times 5 \times 1)$\\

\citet{yepes2020grasp} & Machine setup & Constant & Variable & Single resource (personnel) & Single period & - & MILP & $(6 \times 5 \times 2)$\\

\citet{fanjul2020models} & Machine setup / Job processing & Constant & Variable & Three resources & Single period & - & MILP + CP & $(400 \times 8 \times 40)$\\

\citet{bitar2021unrelated} & Job processing & Constant & Fixed (one per job) & Single resource & Single period & Machine eligibility & MILP & $(50 \times 3 \times 6)$\\


\citet{yunusoglu2022constraint} & Job processing & Constant & Variable & Multiple resources & Single period & Release date, machine eligibility, and job precedence constraint & CP & $(60 \times 8 \times 2)$\\

\citet{chen2022unrelated} & Job processing & Constant & Fixed (one per job) & Single resource (reticle) & Single period & Release date & MILP & $(20 \times 3 \times 5)$\\

\citet{avgerinos2023scheduling} & Machine setup & Constant & Fixed (one per job) & Single resource & Single period & Job splitting & Heuristics + MILP + CP & $(200 \times 20 \times 5)$\\

\textbf{Our work} & Machine setup + Job processing & Variable & Fixed (one per job) & Single resource (personnel) & Multi-period & Release time, delivery time, and machine eligibility & MILP & $(120 \times 8 \times 7)$\\

\bottomrule
\end{tabular}
}
\end{table}

Another group of studies solved UPMS, assuming auxiliary resources are required only during machine setups. \cite{costa2013hybrid} proposed a time-continuous MILP model and a genetic algorithm to address UPMS with personnel allocation as the auxiliary resource. They considered scenarios where the number of available personnel is less than the number of machines and personnel are needed only during machine setups. The MILP model could achieve the global optimum solution for problems with up to 10 jobs, 5 machines, and 3 resources. For problems up to 200 jobs, 20 machines, and 10 resources, the genetic algorithm was employed. \cite{yepes2020grasp} also attempted to address the same problem, presenting a time-indexed MILP model and a novel metaheuristic algorithm, GRASP (Greedy Randomized Adaptive Search Procedure), to tackle the problem. In both works, the MILP model was able to find the optimal solution on small-scale problems with no more than 10 jobs, 5 machines, and 3 total available resources, while metaheuristics could find sub-optimal solutions for problems with 250 jobs and 30 machines in less than a minute. \cite{bektur2019mathematical} developed a MILP model and two metaheuristics (simulated annealing and tabu search) to solve UPMS with a single common server necessary to set up machines. The mathematical model was able to find feasible solutions for problems sizes of 25 jobs and 5 machines when the solution time limit was set to 30 minutes. The metaheuristics achieved feasible solutions for problems with up to 100 jobs and 10 machines. More recently, \cite{avgerinos2023scheduling} proposed a Logic-Based Benders Decomposition (LBBD) approach to solve UPMS with the possibility of splitting jobs and the requirement of personnel to set up machines. The LBBD approach consisted of a master and a sub-problem. In the master problem, a MILP model was employed to solve the assignment of jobs to machines assuming infinite availability of personnel. Then, the sub-problem was formulated as a CP model to optimize the processing sequence of jobs on the machine considering the constraint on the available number of personnel. The authors enhanced the LBBD solving speed by introducing a greedy heuristic algorithm to find an initial solution for the master problem. The proposed method could find near-optimal solutions for problems with 200 jobs, 20 machines, and 5 personnel. Lastly, \cite{fanjul2020models} introduced a comprehensive time-continuous MILP formulation to tackle UPMS with additional resources. Their model accommodated scenarios in which resources were only necessary for job processing, machine set up, or both. This MILP formulation effectively handled problem instances involving up to 50 jobs. Additionally, the authors devised a three-phase decomposition-based algorithm to efficiently address large-scale problems, comprising up to 400 jobs, 8 machines, and 9 resources.\\

\Cref{tab:summaries} provides a summary of key features found in related literature on production scheduling. Prior works addressed the resource-constrained UPMS, considering additional resources for job processing, machine setups, or both. In our work, inspired by a real case study in a food processing plant, we introduce personnel requirements as an auxiliary resource for both supervising machine operations during job processing and setting up machines between successive jobs. Looking at the "Resource availability" and "Scheduling period" columns in \Cref{tab:summaries}, we observe that \textbf{previous works} typically assumed a single-period scheduling horizon where resources are constantly available. However, real-world scenarios may involve variations in resource availability over time. For example, the availability of personnel may fluctuate during different time periods (e.g., weeks) within the scheduling horizon due to factors such as public holidays, vacations, or a company decision to temporarily reduce shift hours in response to low customer demand. Failing to account for these variations in resource availability can result in the generation of production schedules that are impractical to implement in practice. \textbf{Our proposed mathematical formulation} addresses variability in resource availability by incorporating a multi-period scheduling horizon and allowing for constraints on the start and end times of each resource availability within each time period. The model specifically considers a single type of resource, which, in this case study, is the personnel. It assumes a fixed number of resources, such as one personnel, is required per machine. The total number of available personnel, specified as an input to the model, is constrained to be less than the number of machines. This restriction is imposed as a resource constraint, limiting the number of machines that can operate in parallel. Furthermore, our MILP formulation accounts for common manufacturing constraints, including release times, delivery times, and machine eligibility constraints, reflecting real-world complexities.

\section{Problem definition and basic idea} \label{sec: Problem definition}
This section begins with an overview of the scheduling problem addressed in this study, followed by presenting three examples that demonstrate the characteristics of the developed mathematical model. The problem involves a set of $n$ jobs to be processed by a set of $m$ machines. Each job is allocated to one machine, and each machine can process a maximum of one job at a time. Machines cannot be stopped while running, and jobs cannot be preempted during processing. The processing time for each job is machine-dependent, indicating that machines are unrelated. Moreover, setup times are required between consecutive jobs, and these setups are both machine-dependent and job sequence-dependent. There is no precedence relationship among jobs. Some jobs are eligible for processing on a subset of machines, and each job must be processed by one of its eligible machines. A set of $k$ personnel is available and each machine requires the assignment of one personnel to supervise it during setup and job processing. The allocated personnel is retained until the machine completes the processing of the current job and is set up for the next job. The scheduling horizon comprises discrete time periods, such as weeks, and each time period is time-continuous, for instance, expressed in minutes. Workers may be fully or semi-available within each time period, with their availability defined by start and end work times. Jobs may not be available at the beginning of the scheduling horizon, and this is addressed by considering different release times. Similarly, jobs may have customer-set deadlines, represented by different delivery times. The release of a job is defined by a release period and a release time. The release period shows the time period in which the order is released, and the release time determines the time during that period the order is released. Similar definitions hold for the delivery period and delivery time. The objective function is to find an optimal schedule that minimizes the total production time under these operating constraints.\\

\textbf{Basic idea}- In the remainder of this section, we present three examples to illustrate the concept behind the mathematical model developed and the main features of the model. For simplicity, these examples assume that each job is eligible for processing on only one machine. However, the same ideas apply to cases without machine eligibility constraints. Time periods in these examples represent weeks and one personnel is required per machine to supervise machine setup and job processing. Workers operate for 7.5 hours per day, five days a week, totaling 2,250 minutes of weekly availability.\\

\begin{figure}[H]
    \centering
    \includegraphics[width=15cm]{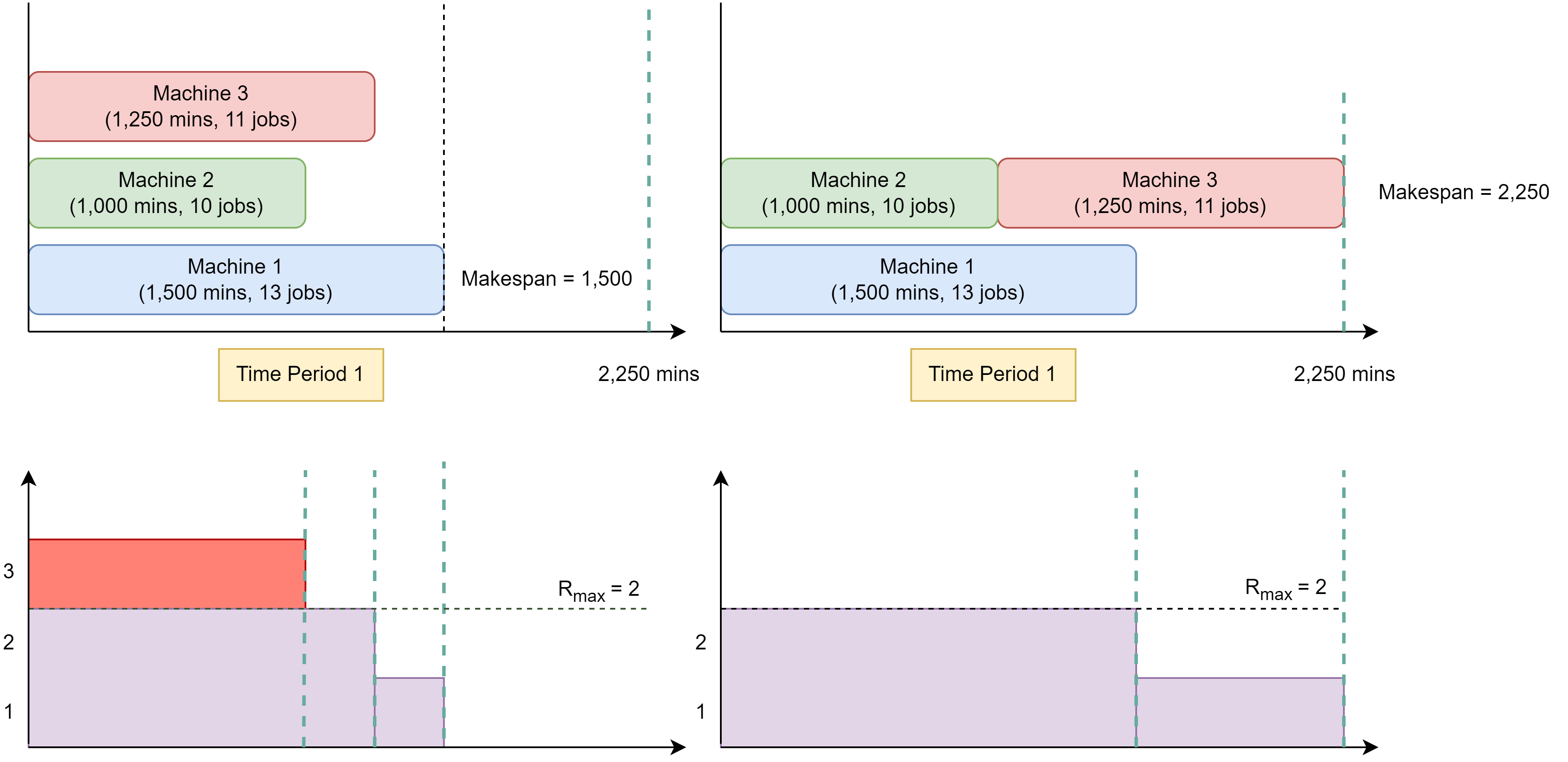}
    \caption{Example of UPMS with personnel availability as a resource constraint}
    \label{fig:ResourceExp}
\end{figure}

\textit{Example 1:} \Cref{fig:ResourceExp} illustrates a production environment with three machines, each processing jobs eligible for production. Each machine requires a personnel to supervise during setup times and job processing. The available number of personnel is limited to two, indicating that only two of the three machines can operate simultaneously. On the left-hand side, the personnel availability constraint is violated, as all three machines run in parallel, requiring three personnel, which exceeds the available number. This violation introduces errors in calculating makespan, starting time, and ending time of jobs, causing the generated production schedule to be infeasible for implementation. Conversely, the right-hand side of \Cref{fig:ResourceExp} adheres to the personnel availability constraint. Consequently, a maximum of two machines operate in parallel. The first personnel is dedicated to Machine 1, while the second personnel completes processing jobs on Machine 1 before moving to Machine 3. As a result, the makespan increases from 1,500 minutes in the infeasible scenario to 2,250 minutes in the feasible scenario.\\

\begin{figure}[H]
    \centering
    \includegraphics[width=15cm]{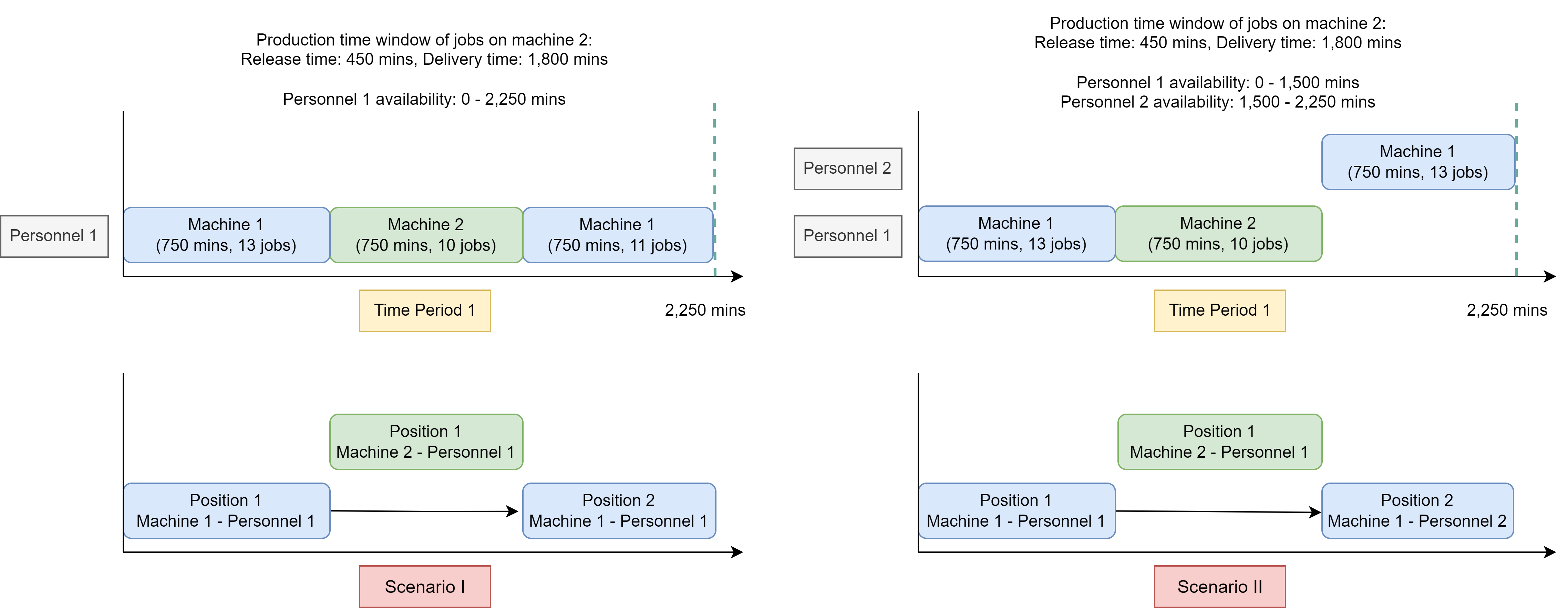}
    \caption{Example of "position" concept introduced to the mathematical model}
    \label{fig:Positions}
\end{figure}

\textit{Example 2:} \Cref{fig:Positions} illustrates the two possible scenarios that require the plant to run a machine more than once in a single time period. In Scenario I, depicted on the left-hand side, there is only one personnel available who should supervise the two given machines. The jobs eligible for processing on Machine 2 have a common production time window. They are released after 450 minutes and should complete processing before 1,800 minutes. Meanwhile, there are 24 jobs eligible for processing on Machine 1. In Scenario I, Machine 1 has to be run twice in this time period. The personnel first processes a portion of jobs belonging to Machine 1, then processes all jobs belonging to Machine 2 while adhering to their production time window. Finally, the personnel runs Machine 1 for the second time to complete processing the remaining jobs belonging to Machine 1. In this scenario, Machine 1 is operated twice, leading to Personnel 1 being assigned to this machine twice. To account for this scenario and other cases where a machine should be run more than once, we introduced the concept of "position". A position resembles a bin in which both a machine and a personnel must be present to process the jobs and set up the machine. The number of positions are defined by the user as an input to the mathematical model. In fact, the position is an index in the model that determines how many times a machine can be run with stoppages in between during a time period. In Scenario I, in \Cref{fig:Positions}, the chart below the production Gantt chart illustrates the positions for the machines. Machine 1 occupies two positions to process the jobs in two separate parts, while Machine 2 utilizes only Position 1 to complete its jobs and does not use Position 2.\\

In Scenario II of \Cref{fig:Positions}, there are two personnel available during the time period, each with partial availability. Personnel 1 is accessible from 0 to 1,500 minutes, while Personnel 2 is available from 1,500 to 2,250 minutes. Moreover, a similar production time window from 450 minutes to 1,800 minutes exists for the jobs processed on Machine 2 as in Scenario I. Similar to Scenario I, Machine 1 needs to operate twice in the time period due to the strict time window of jobs eligible for production on Machine 2. As a result, Machine 1 requires two positions, and Machine 2 needs one position. Different from Scenario I, Position 2 of Machine 1 is occupied by Personnel 2 rather than Personnel 1 due to their distinct work shift schedules. In Scenarios I and II, we assumed that machines can be run two times with a stoppage in between within a time period; therefore, we defined two positions for each. However, depending on the number of possibilities to run a machine within a time period, the number of positions can be adjusted as an input to the model.\\
 
\begin{figure}[H]
    \centering
    \includegraphics[width=15cm]{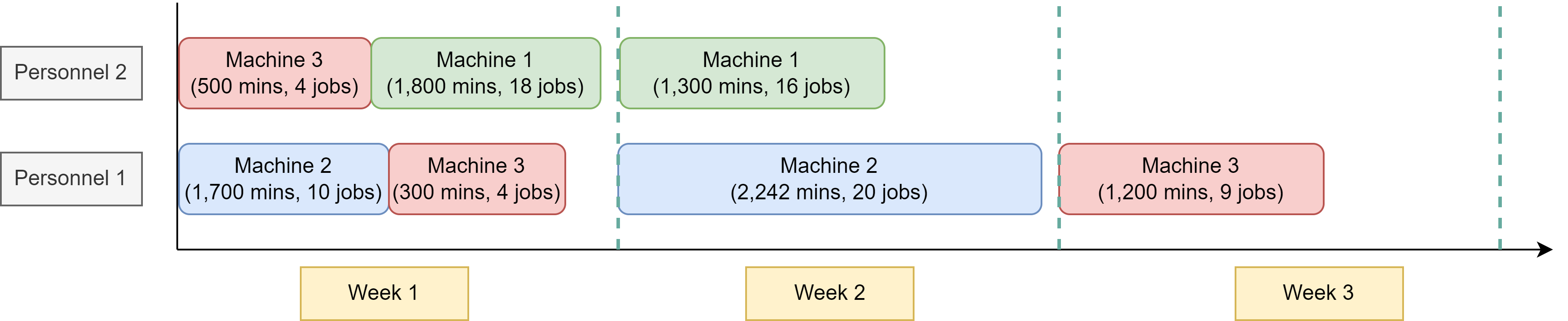}
    \caption{Example of connectivity types in the scheduling problem}
    \label{fig:connectivity}
\end{figure}

\textit{Example 3:} \Cref{fig:connectivity} illustrates three situations in which ensuring connectivity between machine operations is essential, either within a time period or spanning two different time periods. In the first situation, Machine 3 in Week 1 operates twice. Considering the setup time between successive jobs, it is essential to ensure the corresponding setup occurs between the last job of the first run of Machine 3 and the first job of the second run of Machine 3 within Week 1. This connectivity constraint is defined through \Cref{eq:t29}, to ensure that the setup takes place between two successive jobs from two separate successive runs of a machine within a week. The second type of connectivity arises when a machine is scheduled to operate in two successive time periods. For instance, in the scenario presented in \Cref{fig:connectivity}, Machine 1 is scheduled to operate in both Week 1 and 2. Therefore, the model must ensure that the setup for Machine 1 is based on the last job processed on Machine 1 in Week 1 and the first job processed on this machine in Week 2. The same connectivity applies to Machine 2 in Week 1 and 2. The third type of connectivity occurs when a machine operates in one time period, skips operation in one or more time periods, and then resumes operation. In \Cref{fig:connectivity}, this type of connectivity is exemplified by Machine 3, which operates in Week 1, pauses in Week 2, and resumes in Week 3. Here again, the model must ensure that the setup time of the machine is based on the last and first job in Week 1 and 3, respectively. The second and third types of connectivity are addressed by adding a balance constraint, as specified in \Cref{eq:t30}.

\section{Mathematical model} \label{sec: MIP} 
We describe the Mixed Integer Linear Programming (MILP) model developed to solve UPMS, considering personnel availability, machine eligibility constraints, release times, and delivery times. To explain the model, we first define the model sets, parameters, and decision variables, as listed in \Cref{tab:nomenclature}. Next, we outline the model objective function and constraints. Model constraints can be divided into two groups. The first group of constraints ensures the allocation, sequencing, and timing of processing jobs on the machines. The second group ensures the same concepts but for the personnel assigned to the machines.

\begin{longtable}[h]{p{0.3\textwidth}p{0.6\textwidth}}
    \caption {An overview of sets, parameters, and decision variables of the proposed model}
    \label {tab:nomenclature}
\\ \hline
\textbf{Sets} &   \textbf{} \\ \hline

$i, j \in {N} = \{1, \ldots, n\}$                    & Set of all jobs\\

$h \in {N_0}, N_0 = N \cup \{0\}$                                 & Set of all jobs plus dummy job, denoted by $0$\\

$l    \in {M} = \{1, \ldots, m\}$                    & Set of all machines\\

$k \in {K} = \{1, \ldots, k\}$                       & Set of all personnel\\

$p, u \in {P} = \{1, \ldots, p\}$                    & Set of all positions\\

$t \in {T} = \{1, \ldots, t\}$                       & Set of time periods\\

$v \in {T_0}, T_0 = T \cup \{0\}$                                 & Set of all time periods plus dummy period, denoted by $0$\\

$\Gamma_l \subset N$                                 & Subset of jobs which can be processed on machine $l$\\

$\Delta_l \subset P$                                 & Subset of positions, belonging to machine $l$\\

$\alpha_l \subset P$                                 & Subset of positions, referring to the first position of each machine $l$\\

\hline
\textbf{Parameters} & \textbf{}  \\ \hline

$P_{il}$           & Processing time of job $i$ on machine $l$, zero if the machine is not eligible to process the job\\
$S_{ijl}$          & Setup time required before processing job $j$ on machine $l$ preceded by job $i$\\
$S_{il}^I$          & Initial setup time before the first job $i$ in the sequence on machine $l$\\
$AVB_{t}^P$         & Available time during time period $t$\\
$Start_{kt}^P$         & Availability of personnel $k$, the start of working time of personnel $k$ during time period $t$\\
$End_{kt}^P$         & Availability of personnel $k$, the end of working time of personnel $k$ during time period $t$\\
$RP_{i} \in {T}$        & Release period of job $i$ in which the job is ready to get scheduled\\
$DP_{i} \in {T}$            & Delivery period of job $i$ in which the job has to be completed for delivery to the customer\\
$RT_{it}$    & Release time of job $i$ during time period $t$, zero if $t \neq RP_{i}$ \\
$DT_{it}$    & Delivery time of job $i$ during time period $t$, equal to $AVB_{t}^P$ if $t \leq DP_{i}$ and zero if $t \geq DP_{i}$\\

\hline
\endfirsthead
\hline

\textbf{Decision variables} &  \\ \hline

${Alc}_{ilpt|p \in \Delta_l}^J$          & 1 if job $i$ is processed on machine $l$ within position $p$ in time period $t$, zero otherwise\\
${Seq}_{ijlpt|i \neq j \& p \in \Delta_l}^J$         & 1 if job $j$ is the successor of job $i$ on machine $l$ within position $p$ in time period $t$, zero otherwise\\
${Beg}_{ilpt}^J$          & 1 if job $i$ is the first job processed on machine $l$ within position $p$ in time period $t$, zero otherwise\\
${End}_{ihlpt}^J$          & 1 if job $i$ is the last job processed on machine $l$ within position $p$ and job $h$ is the next job processed in the next position in time period $t$, zero otherwise\\
${ST}_{ilpt}^J \geq 0$    & Non-negative continuous variable, the start time of job $i$ processed on machine $l$ within position $p$ in time period $t$\\
${EN}_{ilpt}^J \geq 0$   & Non-negative continuous variable, the end time of job $i$ processed on machine $l$ within position $p$ in time period $t$\\

${AuxVar}_{hlv}$          & Auxiliary variable to keep connection between production sequence of jobs in time period $v$ and $v+2$ on machine $l$ when there is no  job scheduled for processing on machine $l$ in time period $v+1$; 1 if job $h$ is the first job processed on machine $l$ in time period $v+2$, zero otherwise\\

${Alc}_{kpt}^P$          & 1 if personnel $k$ is allocated to position $p$ in time period $t$, zero otherwise\\
${Seq}_{kput}^P$         & 1 if personnel $k$ is allocated to position $p$ before position $u$ in time period $t$, zero otherwise\\
${Beg}_{kpt}^P$          & 1 if personnel $k$ begins its work by being allocated to position $p$ in time period $t$, zero otherwise\\
${End}_{kpt}^P$          & 1 if personnel $k$ ends its work by being allocated to position $p$ in time period $t$, zero otherwise\\
${ST}_{kpt}^P \geq 0$    & Non-negative continuous variable, the scheduled start time of personnel $k$ in position $p$ in time period $t$\\
${EN}_{kpt}^P \geq 0$    & Non-negative continuous variable, the scheduled end time of personnel $k$ in position $p$ in time period $t$\\

\end{longtable}

\textbf{Objective function:}

Objective function given in \Cref{eq:t1} minimizes the total production time over the scheduling horizon. 


\begin{equation}\label{eq:t1}
\begin{aligned}
\text{minimize} \ Z = & \sum_{i \in N, l \in M, p \in P, t \in T} {P}_{il} {Alc}_{ilpt}^J  + \sum_{i,j \in N, l \in M, p \in P, t \in T} {S}_{ijl} {Seq}_{ijlpt}^J \\
& + \sum_{i \in N, l \in M,  p \in P, t \in T} \left(S_{il}^I Beg_{ilpt}^J + \sum_{j \in N} {S}_{ijl} {End}_{ijlpt}^J \right)
\end{aligned}
\end{equation}


\textbf{Constraints:} The constraints can be divided to those related to job scheduling and personnel scheduling.\\

\textbf{Job scheduling constraints}



\textit{Job allocation}: \Cref{eq:t4} guarantees that each job is executed exactly once on one of the eligible machines. \Cref{eq:4} specifies that the allocation of jobs that are non-eligible for production on machine $l$ is set to zero.

\begin{equation}\label{eq:t4}
\sum_{l \in {M}, p \in {P}, t \in {T}} Alc_{ilpt}^J = 1, \quad \forall i \in {N}
\end{equation}

\begin{equation}\label{eq:4}
\sum_{i \notin \Gamma_l, p \in {P}, t \in {T}} Alc_{ilpt}^J = 0, \quad \forall l \in {M}
\end{equation}

\textit{Starting and ending time}: For every job $i$, the difference between its starting and ending time must be at least equal to the processing time of job $i$, \Cref{eq:t5}. The ending time of job $i$ can be nonzero only if the job is allocated to a machine, \Cref{eq:t6}.

\begin{equation}\label{eq:t5}
EN_{ilpt}^J - ST_{ilpt}^J \geq {P}_{il} \ Alc_{ilpt}, \quad \forall i \in N, l \in {M}, p \in \Delta_l, t \in {T}
\end{equation}

\begin{equation}\label{eq:t6}
EN_{ilpt}^J \leq AVB_{t}^P \ Alc_{ilpt}, \quad \forall i \in {N},l \in {M}, p \in \Delta_l, t \in {T}
\end{equation}

\textit{No overlapping production sequence}: The starting time of job $j$ occurs after the ending time of job $i$ if job $j$ is processed after job $i$ on the same machine and position as job $i$.

\begin{equation}
\begin{gathered}\label{eq:t7}
ST_{jlpt}^J - EN_{ilpt}^J \geq AVB_{t}^P \left(Seq_{ijlpt}^J - Alc_{ilpt}^J \right) + {ST}_{ijl} {Seq}_{ijlpt}^J, \\
\quad \forall l \in {M}, p \in \Delta_l, t \in {T}, {i,j} \in N, i \neq j
\end{gathered}
\end{equation}

\textit{Begin and end constraints}: When sequencing jobs on position $p$, there is only one job at the beginning (\Cref{eq:t8}) and one at the endpoint (\Cref{eq:t9}) of the sequence.

\begin{equation}\label{eq:t8}
\sum_{i\in{N}} Beg_{ilpt}^J \leq 1, \quad \forall l\in{M}, p\in \Delta_l, t\in{T}
\end{equation}

\begin{equation}\label{eq:t9}
\sum_{i\in{N}, h\in{N_0}} End_{ihlpt}^J \leq 1, \quad \forall l\in{M}, p\in \Delta_l, t\in{T}
\end{equation}

\textit{Precedence and successive constraints}: \Cref{eq:t10,eq:t11} are specified to guarantee that every job has one predecessor on the position in which it is processed. The constraint given in \Cref{eq:t10} ensures the allocated job $i$ to machine $l$ is preceded only by one other job $j$. If job $i$ is not the initial event, there must be an existing job $j$, occurring before job $i$.

\begin{equation}\label{eq:t10}
Beg_{ilpt}^J + \sum_{j | i \neq j} Seq_{jilpt}^J = Alc_{ilpt}^J, \quad \forall 
i\in{N}, l\in{M}, p\in \Delta_l, t\in{T}
\end{equation}

The constraint given 
 in \Cref{eq:t11} ensures that the job $i$ allocated to position $p$ is succeeded only by one other job $j$. If job $i$ is not the last job, there must be an existing job $j$, occurring after job $i$.

\begin{equation}\label{eq:t11}
\sum_{h\in{N_0}} End_{ihlpt}^J + \sum_{j | i \neq j} Seq_{ijlpt}^J = Alc_{ilpt}^J, \quad \forall 
i\in{N}, l\in{M}, p\in \Delta_l, t\in{T}
\end{equation}

\textit{Release and delivery constraints}: \Cref{eq:t13,eq:t14,eq:t15,eq:t16} account for possible release and delivery dates specific to each job. To incorporate these constraints into the model, the given release and delivery dates are used to determine the release and delivery time periods ($RP_{i}, DP_{i}$), along with their respective time values ($RT_{it}, DT_{it}$). For instance, when considering job $i$ released at time $\tau$ within the second time period in the scheduling horizon, its release period is denoted as ${RP}_i=2$ and its release time as ${RT}_{i,t=2}=\tau$.


\begin{equation}\label{eq:t13}
\sum_{l \in {M}, p \in \Delta_l} RP_{i} Alc_{ilpt}^J \leq \sum_{l \in {M}, p \in \Delta_l} Alc_{ilpt}^J |t|, \quad \forall i \in N, t \in T
\end{equation}

\begin{equation}\label{eq:t14}
\sum_{l \in {M}, p \in \Delta_l} Alc_{ilpt}^J |t| \leq DP_{i}, \quad \forall i \in N, t \in T
\end{equation}

\begin{equation}\label{eq:t15}
\sum_{l \in {M}, p \in \Delta_l} RT_{i,t} Alc_{ilpt}^J \leq \sum_{l \in {M}, p \in \Delta_l, t \in {T}} ST_{ilpt}^J, \quad \forall i \in N
\end{equation}

\begin{equation}\label{eq:t16}
\sum_{l \in {M}, p \in \Delta_l, t \in {T}} EN_{ilpt}^J \leq \sum_{l \in {T}, p \in \Delta_l, t \in {T}} DT_{it}^J, \quad \forall i \in N
\end{equation}

\textbf{Personnel scheduling constraints}

\textit{Personnel allocation}: \Cref{eq:t17} ensures that personnel $k$ is allocated to position $p$ when there exist a beginning job for this position to be processed by machine $l$ in time period $t$.

\begin{equation}\label{eq:t17}
\sum_{ i \in {N}} Beg_{ilpt}^J = \sum_{k \in{K}} Alc_{kpt}^P, \quad \forall l \in {M}, p \in \Delta_l, t \in {T}
\end{equation}

\textit{Personnel quantity requirement}: \Cref{eq:t18} ensures the maximum allocation of one personnel to position $p$, for both machine setup and job processing.

\begin{equation}\label{eq:t18}
\sum_{k \in{K}} Alc_{kpt}^P \leq 1, \quad \forall p \in {P}, t \in {T}
\end{equation}

\textit{Position utilization sequence}: \Cref{eq:t19} ensures that the position $p+1$ belonging to the machine $l$ is not activated unless the position $p$ from the same machine is already utilized. This constraint guarantees the utilization of positions belonging to a machine is in a sequential manner.

\begin{equation}\label{eq:t19}
\sum_{k \in{K}} Alc_{kp+1t}^P \leq \sum_{k \in{K}} Alc_{kpt}^P, \quad \forall p,p+1 \in \Delta_l, l \in {M}, t \in {T}
\end{equation}

\textit{Starting and ending times}: For each personnel $k$, the time difference between their start and finish time on position $p$, during time period $t$, must be at least equal to the total time spent processing the jobs assigned to position $p$, as specified in \Cref{eq:t20}. Personnel $k$ can only have a non-zero ending time if they are allocated to a machine, as outlined in \Cref{eq:t21}.

\begin{equation}\label{eq:t20}
\begin{gathered}
EN_{kpt}^P - ST_{kpt}^P \geq \sum_{i \in N} P_{il} {Alc}_{ilpt}^J + \sum_{i,j \in N \, | \, i \neq j} S_{ijl} Seq_{ijlpt}^J \\
+ \sum_{i \in N} \left(S_{il}^I Beg_{ilpt}^J + S_{ihl} \sum_{h \in N} {End}_{ihlpt}^J \right),
\quad \forall l \in M, p \in \Delta_l, t \in T
\end{gathered}
\end{equation}

\begin{equation}\label{eq:t21}
{EN}_{kpt}^P \leq {AVB}_{t}^P {Alc}_{kpt}^P, \quad \forall k \in {K}, p \in {P}, t \in {T}
\end{equation}

\Cref{eq:t22} ensures that for each job $i$, its start time on machine $l$ in position $p$ occurs no earlier than the start time of personnel $k$ responsible for processing the job in position $p$, increased by the initial setup time required before the first job to initiate production on machine $l$.

\begin{equation}\label{eq:t22}
{ST}_{ilpt}^J \geq \sum_{k \in K} {ST}_{kpt}^P + S_{il}^I Beg_{ilpt}^J - {AVB}_{t}^P \left(1 - {Alc}_{ilpt}^J \right), \quad \forall i \in N, l \in {M}, p \in \Delta_l, t \in {T}
\end{equation}

Similarly, \Cref{eq:t23} guarantees that for each job $i$, its finishing time on machine $l$ takes place no later than the scheduled end time of personnel $k$ responsible for processing the job, reduced by the cleaning time needed after the last job, which has completed its processing on machine $l$.

\begin{equation}\label{eq:t23}
{EN}_{ilpt}^J + S_{ihl} \sum_{h \in N_0} {End}_{ihlpt}^J \leq \sum_{k \in K} {EN}_{kpt}^P, \quad \forall {i} \in N, l \in {M}, p \in \Delta_l, t \in {T}
\end{equation}

\textit{No overlapping personnel allocation}: \Cref{eq:t24} guarantees that the starting time of personnel $k$ on position $u$ occurs after the ending time of personnel $k$ on position $p$ if position $p$ precedes position $u$.

\begin{equation}\label{eq:t24}
ST_{kut}^P - EN_{kpt}^P \geq  AVB_{t}^P \left(Seq_{kput}^P - Alc_{kpt}^P \right), \quad \forall k \in {K}, {p,u} \in P, t \in {T}, p \neq u
\end{equation}

\Cref{eq:t25} guarantees the starting time of position $p+1$ always occurs after the ending time of position $p$.

\begin{equation}\label{eq:t25}
\sum_{k \in K} ST_{kp+1t}^P \geq \sum_{k \in K} EN_{kpt}^P + AVB_{t}^P \left( \sum_{k \in K} Alc_{kp+1t}^P - \sum_{k \in K} Alc_{kpt}^P \right), \forall k \in {K}, {p} \in \Delta_l, t \in {T}
\end{equation}

\textit{Begin and end constraints}: Each personnel $k$ begins their work on one of the positions (\Cref{eq:t26}) and also finishes their work on one of the positions (\Cref{eq:t27}). The beginning and ending position for each personnel can be either the same or different.

\begin{equation}\label{eq:t26}
\sum_{p\in{P}} Beg_{kpt}^P \leq 1, \quad \forall k\in{K}, t\in{T}
\end{equation}

\begin{equation}\label{eq:t27}
\sum_{p\in{P}} End_{kpt}^P \leq 1, \quad \forall k\in{K}, t\in{T}
\end{equation}

\textit{Precedence and successive constraints}: Similar to \Cref{eq:t10,eq:t11}, which determines the job sequence, \Cref{eq:t28,eq:t29} are specified to determine the sequence of positions visited by each personnel.

\begin{equation}\label{eq:t28}
Beg_{kpt}^P + \sum_{u | p \neq u} Seq_{kupt}^P = Alc_{kpt}^P, \quad \forall k\in{K}, p\in{P}, t\in{T}
\end{equation}

\begin{equation}\label{eq:t29}
End_{kpt}^P + \sum_{u | p \neq u} Seq_{kput}^P = Alc_{kpt}^P, \quad \forall k\in{K}, p\in{P}, t\in{T}
\end{equation}

\textit{Personnel availability}: \Cref{eq:t30,eq:t31} guarantee that the start and end work time of personnel $k$ within position $p$ in time period $t$ remains within the respective available time window for that personnel and position.

\begin{equation}\label{eq:t30}
Alc_{kpt}^P Start_{kpt}^P \leq ST_{kpt}^P, \quad \forall k \in K, p \in P, t \in T
\end{equation}

\begin{equation}\label{eq:t31}
EN_{kpt}^P \leq End_{kpt}^P, \quad \forall k \in K, p \in P, t \in T
\end{equation}

\textit{Connectivity constraints}: \Cref{eq:t32} maintains the connectivity between the production sequence of position $p$ and position $p+1$ that belong to machine $l$ within the time period $t$. \Cref{eq:t33} eliminates creating any closed loop in the production sequence of position $p$.   

\begin{equation}\label{eq:t32}
\begin{gathered}
Beg_{hlp+1t}^J \geq \sum_{i \in N} End_{ihlpt}^J + \left(\sum_{k \in K} {Alc}_{kp+1t}^P - \sum_{k \in K} {Alc}_{kpt}^P \right), \\
\quad \forall h \in N_0, l \in {M}, p,p+1 \in \Delta_l, t \in T
\end{gathered}
\end{equation}

\Cref{eq:t33} is the balance constraint, maintaining connectivity between the production sequence of machine $l$ from time period $t$ to time period $t+1$. The auxiliary variable ${AuxVar}_{hlt}$ becomes active whenever the model decides to not schedule any job on machine $l$ during time period $t$. Whenever ${AuxVar}_{hlt}$ becomes 1, it transfers the next job $h$ to the time period $t+2$.

\begin{equation}\label{eq:t33}
\begin{gathered}
\sum_{i \in N, p \in \Delta_l} End_{ihlpt}^J - \sum_{p,p+1 \in \Delta_l} Beg_{hlp+1t}^J +{AuxVar}_{hlt-1} = Beg_{hl\alpha_l t+1}^J + {AuxVar}_{hlt}, \\
\quad \forall h \in N_0, l \in {M}, t < |T|
\end{gathered}
\end{equation}






\subsection{Two-step solution approach} \label{sec: TSSA}
Solving UPMS with personnel requirements as an additional resource can be carried out in two steps. The first step involves finding an initial feasible solution by allocating jobs to the machines, ensuring compliance with the given constraints, including personnel availability, machine eligibility, and production time windows. In the second step, the initial feasible solution is optimized by altering the processing sequence of jobs while still adhering to all problem constraints. A similar approach has been employed by \cite{tran2016decomposition} and \cite{fanjul2019reformulations} to solve conventional UPMS, and by \cite{avgerinos2023scheduling} to address UPMS with personnel availability constraints. In these studies, the authors divided the problem into job allocation and job sequencing problems. In our study, we develop a Two-Step Solution Approach (TSSA) to solve UPMS with a personnel availability constraint. In the first step, \Cref{eq:t4} is relaxed to \Cref{eq:t35}, and the objective function is changed from \Cref{eq:t1} to \Cref{eq:t36}, with all other constraints remaining unchanged. The primary goal of the first step is to maximize the number of jobs accepted while meeting all constraints except \Cref{eq:t4}. The first step continues until the optimality gap becomes zero or the time limit is reached. If the optimum solution with zero gap is found, the model can achieve either of two cases: \textit{Case 1}: All jobs are successfully scheduled for processing, indicating that the problem is feasible. The solution from this case serves as a warm-start for the second step. \textit{Case 2}: Only a subset of jobs is scheduled for processing, revealing the inherent infeasibility. In this situation, all jobs cannot be scheduled within the given constraints and scheduling horizon. When the first step recognizes infeasibility, it terminates early without starting the second step, shortening the overall runtime. In the feasible case (Case 1), the second step uses the solution obtained in the first step as a warm-start, reducing the time spent finding the initial feasible solution. This second step aims to minimize the total production time, as defined in \Cref{eq:t1}, by altering the processing sequence of jobs, subject to all constraints (\Cref{eq:t4} - \Cref{eq:t33}).


\begin{equation}\label{eq:t35}
\sum_{l \in {M}, p \in {P}, t \in {T}} Alc_{ilpt}^J \leq 1, \quad \forall i \in {N}
\end{equation}

\begin{equation}\label{eq:t36}
{max}\ Z_1= \sum_{i \in {N},l \in {M}, p \in {P}, t \in {T}} Alc_{ilpt}^J
\end{equation} 

\section{Computational experiments} \label{sec: computational experiments}
Computational experiments were conducted by applying the developed mathematical model to both synthetic problem instances and a real-world case study dataset. The use of synthetically generated instances allows for an analysis of the model performance in a general context, while the real-world case study serves to demonstrate the practical application of the model and highlight potential managerial insights that can be derived from its implementation.

\subsection{Instance generation and experimental setting}
\label{sec: Instance generation}
To evaluate the versatility and robustness of the developed model, a set of randomly generated instances is created. The developed data generation software is made publicly available at our GitHub repository \url{https://github.com/tcharte/Machine-Scheduling-Data-Generator} (Accessed on 15 January 2024). The instances are generated in Python with input parameters for the number of jobs \( J \in \{30, 50, 60, 75, 90, 100, 120\} \), the number of machines \( M \in \{2, 3, 4, 5, 6, 7, 8\} \), the number of time periods (scheduling horizon) \( P \in \{1, 2, 3\} \) (in weeks), and personnel capacity \( K \in \{1, 2, 3, 4, 5, 6, 7\} \). The code imposes constraints on these parameters to ensure the generated instances represent plausible and challenging scenarios. To adhere to the specifications laid out in \cite{arnaout2010two, tran2016decomposition, fanjul2019reformulations}, the ratio of jobs to the product of machines and weeks is constrained to be not less than 15. This condition reflects a common scenario in a busy production settings where machines are highly utilized, averaging at least 15 jobs per machine. To consider personnel availability as a resource constraint, their available number is less than the number of machines. Also, the ratio of machines to personnel is less than or equal to three, aligning with practical manufacturing constraints. Furthermore, the scheduling instances are created with diverse scheduling scenarios by generating instances that span all combinations of the previously mentioned parameters and all combinations of including or excluding production time windows and machine eligibility constraints. This encompasses instances with no time windows or machine eligibility constraints, with time windows but no machine eligibility constraints, without time windows but with machine eligibility constraints, and with both time windows and machine eligibility constraints. In instances where time windows are implemented, each job is assigned its unique release time and delivery time. This assumption facilitates the evaluation of the mathematical model performance when confronted with extra constraints imposed by the time windows. Additionally, in cases where machine eligibility constraints are enforced, it is assumed that only one machine from the pool of available machines is qualified to handle a specific job and jobs are evenly distributed among the machines. Therefore, each machine will be eligible for processing a similar number of jobs.\\

Job processing and setup times are determined using a discrete uniform probability distribution, ensuring randomness in the generated instances. To establish the lower and upper bounds of these probability distributions, we first compute the mean value, representing the average time required for both processing a job and setting up the machine for the subsequent task. This mean value is calculated under the assumption of an equal distribution of jobs among the machines. If we denote this value as the "Average number of jobs per machine," each machine processes this number of jobs and requires setups equal to this value minus one. The mean value is derived using \Cref{eq:mean}, where the total available time is divided by the product of the number of machines and twice the average number of jobs allocated to each machine minus one. Subsequently, the upper and lower bounds of the job processing and machine setup times are set to \( \frac{4}{3} \) and \( \frac{2}{3} \) of the mean, respectively.

\begin{equation}\label{eq:mean}
\text{Mean value} = \frac{\text{Total available time}}{\text{No. of machines} \times (2 \times \text{Avg. No. of jobs allocated to each machine} - 1)}
\end{equation}

The total available time in \Cref{eq:mean} is computed based on the number of personnel available, the number of time periods in the scheduling horizon, and the availability of personnel within each time period. The time periods are assumed to represent weeks within the scheduling horizon to reflect common practices in manufacturing environments. Additionally, each week comprises five workdays, with personnel assumed to work 7.5 hours per day, resulting in 2,250 minutes of weekly time availability per personnel. All processing times are randomly generated within the calculated lower and upper bounds, varying depending on the machine to achieve machine-dependent processing times. Initial setup times are uniformly set as the upper bound. Subsequently, sequence-dependent and machine dependant setup times are generated within the range of upper and lower bounds using a coordinate method that ensures the triangle inequality is met. This method involves generating random coordinates for every two jobs on every machine, calculating distances between the two coordinates, and using these distances to determine the setup times as described in \cite{tran2016decomposition}. The setup times are non-symmetric (i.e., setup time from job A to B may differ from the setup time from B to A). This asymmetry captures the realistic scenario where setup times can vary depending on the job sequence.

\subsection{Case study} \label{sec: case study}
The case study in this research involves a food processing plant dedicated to producing over 100 different food powders. This case study serves as a real-world application of the mathematical model outlined in \Cref{sec: MIP}. The plant operates under a batch production policy, and orders encompass a combination of Make-To-Stock (MTS) and Make-To-Order (MTO) products. For MTS products, planned orders are calculated based on factors such as the current stock position, safety stock, demand forecast, and previously placed orders. On the other hand, planned orders for MTO products are specified by customers. Eventually, for both MTS and MTO products, the number of batches for production is determined by dividing the planned order amount by the standard order quantity. In line with the plant operational policy, batches of the same product with a common release and delivery time are aggregated and scheduled for production as a single job. This practice, which aims to reduce the time required to clean the machine between batches, improves overall efficiency. The production due date (delivery time) of jobs, whether MTS or MTO products, is dictated by customer requests. Similarly, their release time is determined on the basis of the available inventory of raw materials necessary to process the job.\\

The shop floor is comprised of three production lines, each dedicated to producing specific types of products, introducing machine eligibility constraints. Each production line consists of five stages: Pre-scaling, bag chute, mixing, sieving, and packing. Notably, the packing stage in each line is identified as the bottleneck due to its longer processing time compared to the other stages. Once the production schedule for jobs in the packing stage is determined, the preceding stages can follow the same schedule. This characteristic transforms the production scheduling into a UPMS scenario, where the packing lines serve as unrelated parallel machines with machine eligibility constraints. Importantly, personnel availability is only sufficient to run two packing lines per shift, making manpower availability a critical resource constraint that must be considered in the schedule to ensure a feasible production plan. The availability of personnel during work shifts is sufficient to run two of the three packing machines, representing a resource constraint that must be considered in the schedule to generate a feasible production plan. Additionally, the plant policy is to run each production line only once a week to minimize availability losses. Therefore, if a personnel is assigned to two lines in a week, they must first complete production on one line before starting production on the other. Three main time components are associated with processing jobs: processing time for packing a job (assumed to be the maximum of personnel and machine time), setup time required before running the packing stage when switching between two jobs from different products, and changeover times. The setup time varies between 10 and 20 minutes, depending on the production line. Changeover times occur to clean and re-setup the machine after processing each batch, with two types: "to a different product" (varying between 25 and 90 minutes, depending on the product and line) and "to the same product" (averaging 7-8 minutes between every two batches). In this study, the summation of setup times and changeover times is considered as sequence-dependent and machine-dependent setup times required for changing production from one batch to the next.\\

\textbf{Potential areas of improvement -} According to the Overall Equipment Effectiveness (OEE) report and discussions with the plant manager, the primary source of downtime and loss analysis is attributed to changeovers (setup times), contributing to one-third of total availability losses. It is crucial to determine the optimal sequence for job production to minimize downtime. Additionally, leveraging campaign manufacturing, which involves producing jobs from the same product in sequence to minimize "to different product changeovers," is a significant solution to reduce total changeover times. Minimizing changeover times has several advantages, including reducing production time. This reduction not only provides extra time but also leads to cost savings and decreased electricity consumption. The additional time gained can be utilized to increase production capacity or allocate for activities such as deep cleaning and line maintenance. Based on discussions with the plant production planner, optimizing the schedule to allow for deep cleans or maintenance during regular workdays instead of weekends (overtime) is an effective strategy to reduce operational costs and enhance staff satisfaction. Implementing a proper scheduling tool, which can accommodate variability in the personnel shift hours, assists the scheduler in evaluating the feasibility of accepting staff vacations based on the weekly hours required to complete the production schedule for that week. In summary, the primary area of improvement for the plant is to decrease production time by minimizing changeover times while adhering to manufacturing constraints such as job release times, delivery times, machine eligibility, and personnel shift schedules.

\section{Results and discussion} \label{sec: Results}
We conducted experiments on the synthetically generated problem instances and case study to assess the model's performance and outcomes. The MILP model was executed on the \cite{AIMMS} using \cite{Gurobi} solver on a single machine to maintain consistency for the computational experiments. The computational experiments were conducted on a single machine with the following specifications: Windows 10 64-bit operating system, an AMD Ryzen 9 3900X processor (12 cores) @ 3.8 GHz processor, and 128 GB of RAM. The analyses presented in this section were performed using \cite{Tableau} under an academic license. By employing synthetic problem instances and the case study, we conducted a comprehensive analysis of managerial implications in the scheduling problem context. This analysis provides valuable insights for decision-makers.

\subsection{Synthetically generated data results} \label{sec:computational}

\begin{figure}[H]
    \centering
    \includegraphics[width=15cm]{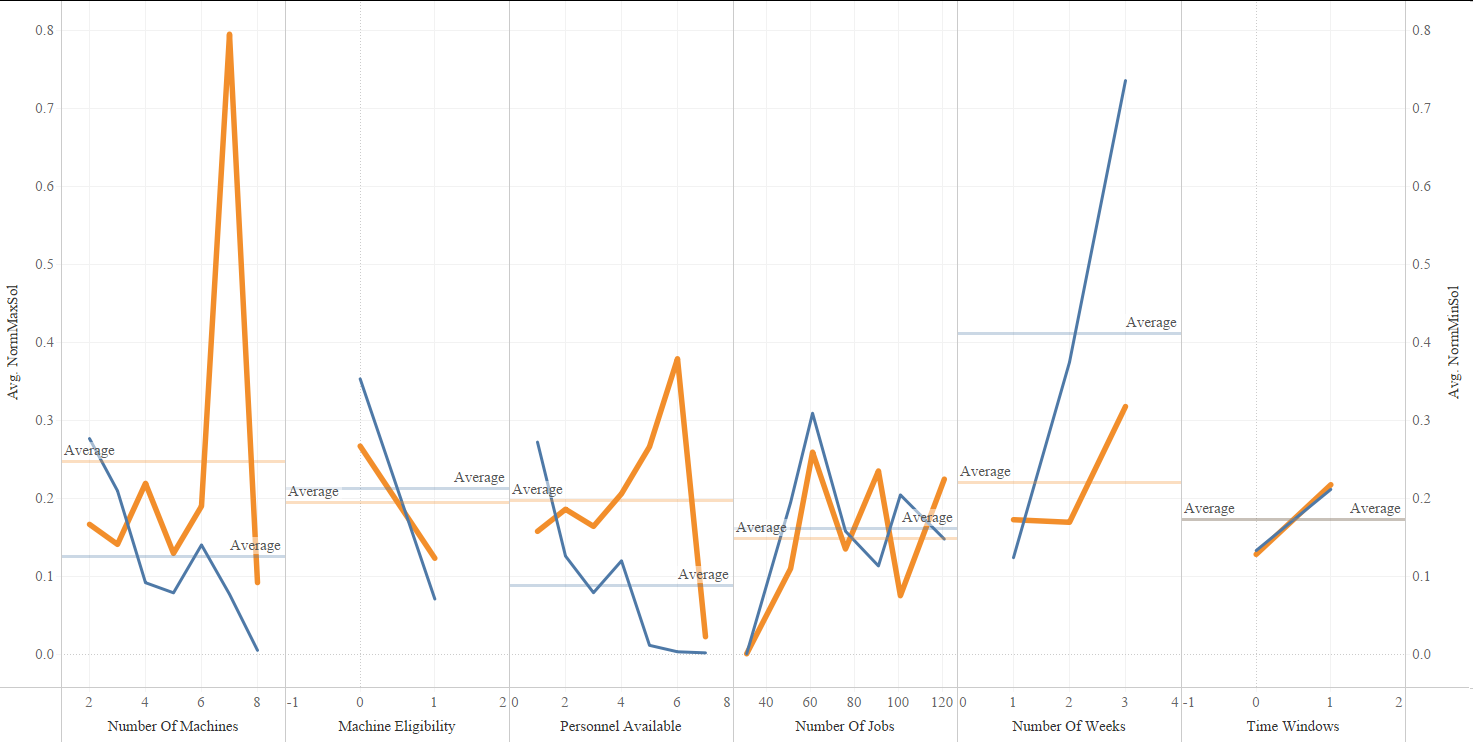}
    \caption{First and second step normalized solution time of model considering influential factors (the solution times of first and second step are represented by the orange and blue line, respectively)}
    \label{fig:SolutionTime}
\end{figure}

As previously discussed, synthetic instances were generated to assess the performance of the developed mathematical model, taking into account various factors that influence production scheduling. These factors encompass the number of jobs, machines, personnel, machine eligibility, and production time window. In this experiment, we generated 118 synthetic problem instances and solved them using the mathematical model. The maximum total solution time was set to 75 minutes, and the acceptable optimality gap was set to 0\% for the first step (maximizing the number of accepted jobs) and 5\% for the second step (minimizing the total production time). The diversity of the problem instances consists of every possible combination of 2 to 10 machines, with 30, 50, 60, 75, 90, 100, and 120 jobs. Additionally, we considered one to three weeks scheduling horizons, and one to seven available personnel in this set of instances. Some conditions, as discussed in \Cref{sec: Instance generation}, were applied to generating the problem instances. Due to adoption of maximum solution time of 45 minutes for the first step and 30 minutes for the second step,we found it necessary to normalize the solution time of each step based on their respective maximum to compare the solution time of these two steps. \Cref{fig:SolutionTime} illustrates the normalized average solution times in the first step (number of accepted job maximization) represented by the orange line and those in the second step (production time minimization) represented by the blue line.

On observing the figure, it becomes evident that obtaining an acceptable solution in the first step consistently requires more time across various factors, such as the number of machines, machine eligibility, personnel, and the number of jobs. This trend suggests that when the solver successfully finds the optimal solution for the problem in the first step, the solution time of optimizing the second step becomes significantly shorter. Furthermore, it is worth noting that determining the lower bound for the model in the first step can be computationally more intensive compared to the second step, consistently requiring more time despite the higher maximum solution time set for the first step. The number of weeks, the number of machines, and the availability of personnel are, respectively, the top three influential factors causing longer solution times, as evidenced by the average computational results.\\

\begin{figure}[H]
    \centering
    \includegraphics[width=15cm]{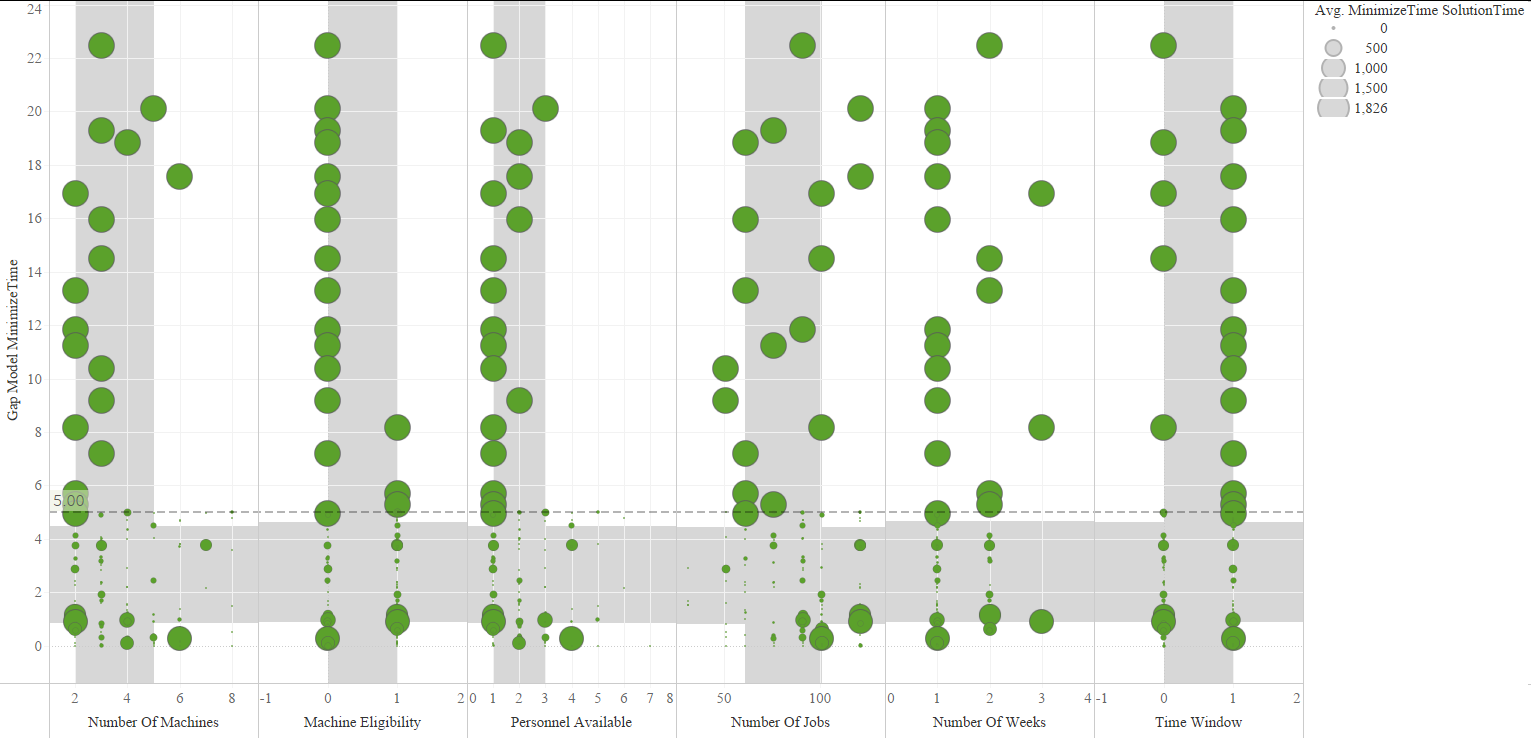}
    \caption{Optimality gap of resolved instances model considering influential factors}
    \label{fig:gap}
\end{figure}

As an additional important component of our analysis, we concentrate on the model gap in addition to solution time. Our goal was to investigate how different factors affected the optimality gap, which is shown in \Cref{fig:gap}. In this figure, we illustrate the gap of resolved instances and represent the solution time of the second step (minimising total production time) proportionally through the depicted green area of circles. It is evident that all points beyond our reference line (5\%) exceeded our termination threshold (i.e., optimality gap of 5\%) and have the maximum possible area (i.e., maximum solution time of 30 minutes). As can be seen from the shaded part, which represents the 95\% confidence interval of the data in each section, over 80\% of the data falls within 5\% optimality gap. Machine eligibility emerges as a significant factor influencing the final gap and the model ability to find optimal solutions. Specifically, the solution time and optimality gap of instances where machine eligibility exists (i.e., represented by value one in \Cref{fig:gap}) significantly are longer and greater, respectively, from problem instances where freedom is given to the model to decide on job-machine allocations. This implies that when machine eligibility does not exist, it takes more time to solve the model but consistently yields solutions with smaller total production time. Therefore, the design and utilization of multi-functional machines for production can substantially reduce total production time. Further details on this phenomenon will be elaborated in \Cref{sec: managerialimplication}. Furthermore, the model encounters difficulties in determining the optimal value when time window constraints are introduced. This leads to extended computational time, and a significant number of cases end before the 5\% gap threshold is reached.

\subsubsection{Managerial implications} \label{sec: managerialimplication}

In this section, the impact of machine eligibility and time window constraints on the production time is analyzed. As it was discussed in \Cref{sec: computational experiments}, the processing and setup times of jobs in the problem instances are dependant on the number of jobs. Consequently, it becomes necessary to compute the average values of processing and setup times before assessing the impact of each factor on them. There is an increase of 54\% in average processing time and 36\% in average setup time across problem instances when machine eligibility constraint exists as compared to when machines are multi-functional. In addition, having the time window constraint adds 38\% and 41\% to the processing and setup time, respectively. This insight gives business owners useful information when comparing the production costs with and without time windows that are determined by each job release time and delivery time. It also emphasizes the importance of production time window since it is adding up to 40\% to the total production time. Therefore, it is important to carefully evaluate the costs associated with meeting release and delivery times in contrast to the potential reduction in total production time achieved by not adhering to the production time windows.



\begin{figure}[H]
    \centering
    \includegraphics[width=15cm]{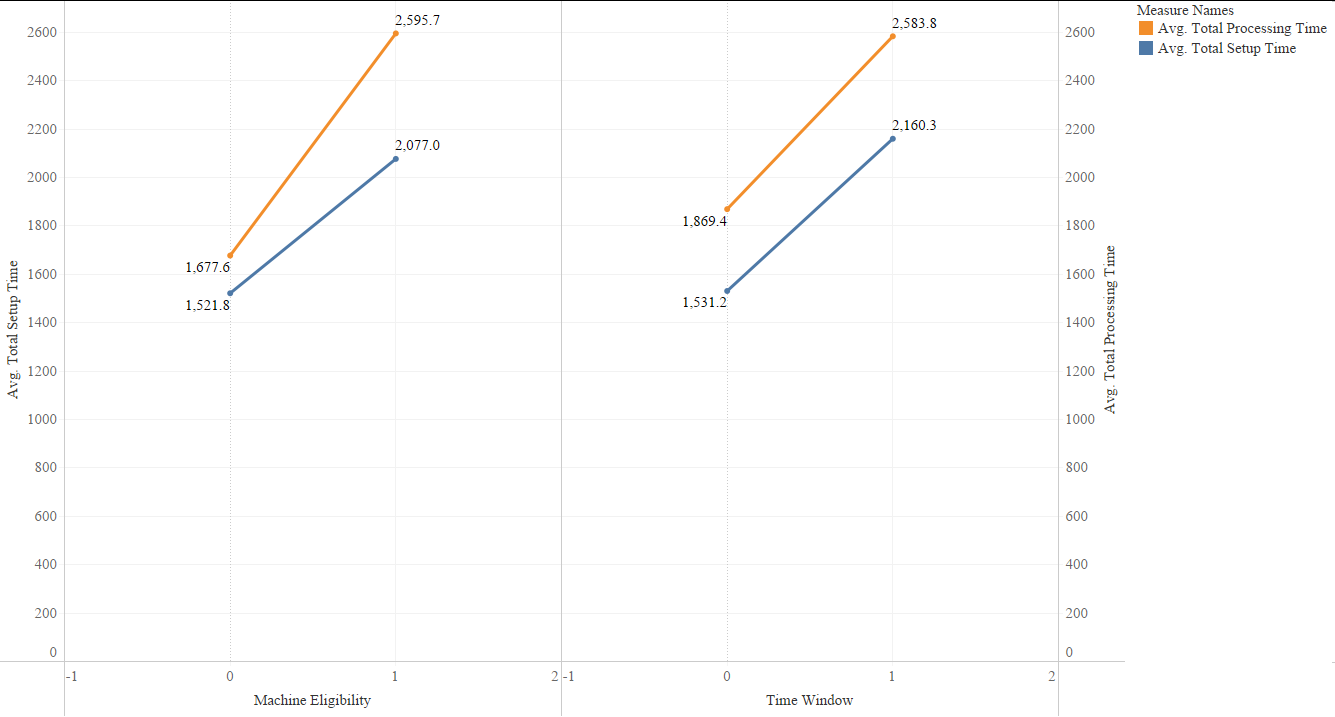}
    \caption{Average total processing time and setup time based on machine eligibility and production time window}
    \label{fig:me-td}
\end{figure}

\begin{figure}[H]
    \centering
    \includegraphics[width=10cm]{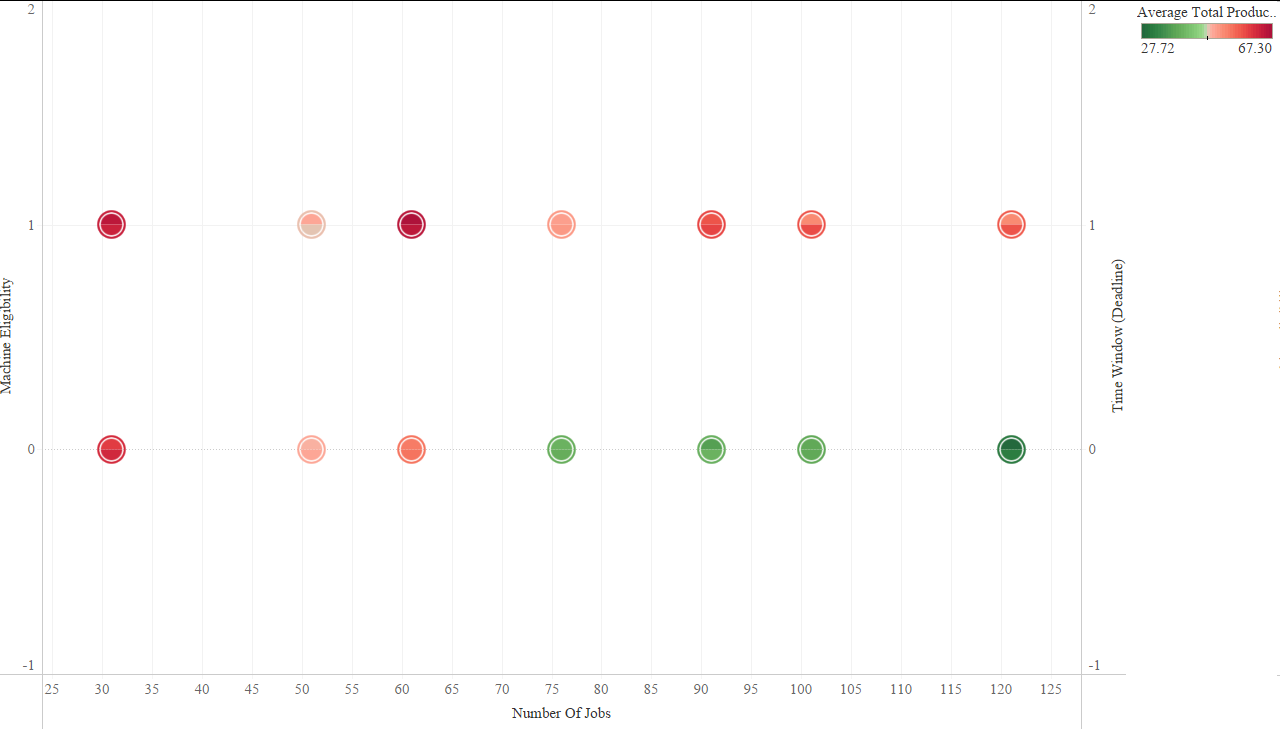}
    \caption{Average total production time per job, categorized by varying machine eligibility and production time window}
    \label{fig:me-td-order}
\end{figure}

We further evaluated the combined contribution of production time window and machine eligibility to the average production time per job. Two semicircles are shown in \Cref{fig:me-td-order}; the lower semicircle shows the effect of time windows on the average production time (processing time plus setup time) per job, and the upper semicircle shows the relative effect of machine eligibility. Looking at \Cref{fig:me-td-order}, one can see that for the problem instances with the low number of jobs, activating and deactivating machine eligibility and time window lead to minimal changes on the average production time per job in the same number of jobs. However, there is a turning point for problem instances with the number of jobs exceeding 75, which indicates a significant effect of machine eligibility and production time window on the average production time per job. This observation implies that machine eligibility and production time window constraints both have a substantial impact on the total production time per job, above a certain job threshold.

\subsection{Case study results}
We showcase the capabilities of the model and its practical utility by applying it to the case study. An overview of the model input parameters for the case study is provided in \Cref{tab:ModelInput}. The planning horizon spans three weeks, aligning with the plant practice of receiving customer approvals for production orders up to three weeks in advance. Each week consists of five workdays, with daily shifts lasting 7.5 hours after accounting for lunch and break times. The plant operates with a total of 90 jobs distributed across three machines, subject to eligibility constraints between jobs and machines. Packing machines 1 and 2 handle the majority of jobs, processing 34 and 48 jobs, respectively, while machine 3 processes only four jobs within the planning horizon. The plant ensures a sufficient supply of raw materials for all planned jobs, allowing the initiation of processing from the first week. This translates to a release time of zero for all jobs. Delivery times vary depending on the weekly deadlines set by the customers. Two personnel are fully available to operate two of the three machines. With the consideration of full personnel availability and weekly deadlines, the company policy dictates running each machine a maximum of once per week. As a result, the weekly allowable personnel-machine assignment is set to one, implying that the number of positions for each machine in each week is set to one. The number of batches within each job ranges from one to 16, dependant on the customer order quantities. The processing time for each job is the aggregate of packing time of batches within a job, "to same product" changeover time  between every two batches, and initial machine setup time.\\

\begin{table}[ht]
\caption{An overview of model input parameters for the case study}
\label{tab:ModelInput}
\resizebox{\columnwidth}{!}{
\centering
\begin{tabular}{m{6.5cm}m{1.5cm}m{4.5cm}m{5cm}}
\toprule
Parameter &  & Parameter & \\
\midrule
Number of jobs & 90 & Weekly time availability & 5 days $\times$ 450 min/day = 2250 min\\ 
Number of machines & 3 & Job batch size & [1, 16]\\
Number of personnel & 2 & Job processing time & [5, 263] min\\
Planning horizon & 3 weeks & Setup time (to different product) & [25, 60] min\\
Weekly allowable personnel-machine assignment & 1 & Setup time (to same product) & 7.2 min\\
\bottomrule
\end{tabular}}
\end{table}

The model results for the case study are presented in \Cref{tab:modelSummary}. The model was executed for two scenarios: Scenario I. When the plant is committed to adhere to customer delivery times and Scenario II. When the plant can deliver customer orders even after the delivery times have passed. The solver maximum run time is set to 1 hour. In both scenarios, the solver efficiently found the optimal solution for the first step in less than 10 seconds, successfully fulfilling all customer orders. In the second step, the optimal solution was found for Scenario I in less than two minutes. However, for Scenario II, the solver reached a feasible solution with a gap of 1.26\% after reaching the maximum run time, i.e., 1 hour. In Scenario I, the solver achieved the optimal solution faster, primarily due to constraints on delivery times, which reduce the solution space. The total production time in Scenario I is 8,941 minutes, equivalent to 19.87 days, assuming 7.5 hours of time availability per day. In Scenario II, the total production time decreases by 3.4 hours. This reduction in production time is attributed to the decrease in setup times. Since the model is allowed to schedule jobs even after their delivery time, the solver can find a better optimal production sequence on each machine, minimizing setup times. The total processing time remains the same in both scenarios, as the model needs to process all the jobs. These results suggest that the model effectively schedules the production of all jobs in both scenarios. However, the total production time is approximately 3.4 hours shorter when the plant is not committed to meeting customer deadlines compared to when deadlines must be adhered to. The saved time can be utilized to accept more jobs, thereby increasing the production volume.\\

\begin{figure}[H]
    \centering
    \includegraphics[width=15cm]{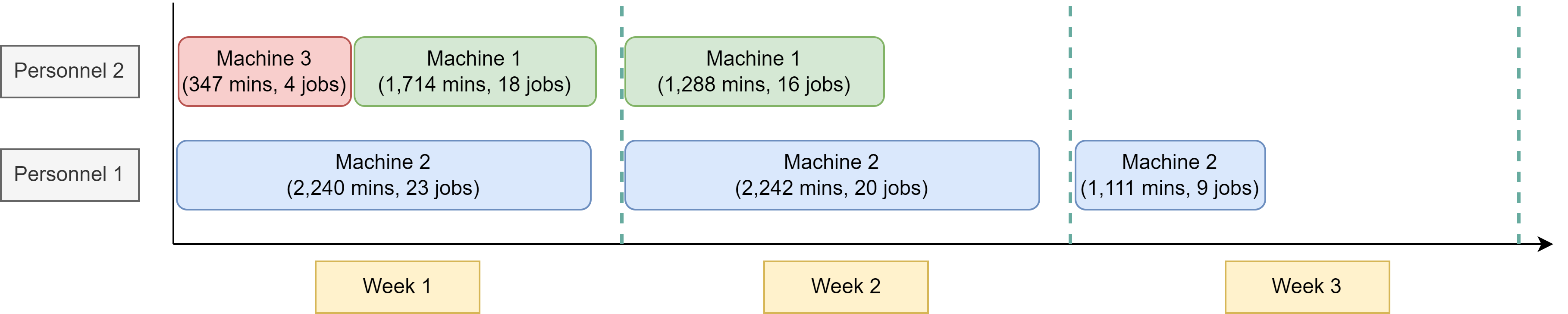}
    \caption{Gantt chart of the production schedule for the case study}
    \label{fig:GanntChart}
\end{figure}

The Gantt chart in \Cref{fig:GanntChart} illustrates the production schedule for Scenario I where deadlines must be met. Personnel 1 is entirely assigned to Machine 2 to process jobs eligible for production on this machine. This machine operates almost continuously for the first two weeks and 2.5 days in the third week. Personnel 2 is assigned to supervise Machines 1 and 3. In the first week, Personnel 2 must first complete jobs on Machine 3 and then start processing jobs belonging to Machine 1. In the second week, Personnel 2 should only run Machine 1 for almost three days to complete the remaining jobs. The weekly utilization level of personnel is visualized in \Cref{fig:utilization}. It is apparent that the shift hours of both personnel are almost fully utilized in the first week. Personnel 1 would be fully utilized in the second week, but will remain unutilized for two and a half days in the third week. Personnel 2 will be idle for 2 days in the second week and fully unutilized in the third week. Based on the level of utilization of personnel in \Cref{fig:utilization} and their weekly time availability, it is evident that two personnel are idle for a combined total of nine days. This is equivalent to almost one week when personnel work for five days each. Similarly, from \Cref{fig:GanntChart}, it can be observed that machines are not fully operational in Weeks 2 and 3. Having information on the weekly production schedule of the machines and the weekly utilization level of personnel during the planning horizon, we can derive the following suggestions. These suggestions aim to optimize resource utilization and enhance operational efficiency during the planning horizon.

\begin{itemize}
    \item \textbf{Increase Production Volume}: The plant can accept additional customer orders to increase production volume on all three machines. Based on \Cref{fig:GanntChart}, orders eligible for production on Machine 1 or 3 can be accepted for production in either Week 2 or Week 3, while additional orders eligible for production on Machine 2 can be only accepted for production in Week 3.

    \item \textbf{Conduct Deep Cleans}: The company can alternatively use the idle time of personnel and machines to perform deep cleanings. Deep cleanings take two days and are necessary to carry out in a timely manner to ensure thorough functionality and productivity of the machines. Worker 2 can use his idle time to conduct a deep clean on Machines 1 or 3 in Week 2 or 3. Either Personnel 1 or 2 can also perform a deep clean on Machine 2 when it completes its production in Week 3.

    \item \textbf{Accommodate Worker Time-Off}: Since personnel would be idle for nine days in total during the planning horizon, the plant can consider personnel requests for taking some days off. This not only allows for a break for the personnel, but also aligns with the observed idle time.
\end{itemize}

\begin{table}[ht]
\caption{The model performance and solution time for the case study}
\label{tab:modelSummary}
\resizebox{\columnwidth}{!}{
\centering

\begin{tabular}{m{4.5cm}m{2.25cm}m{3cm}m{3cm}m{3.25cm}m{3cm}m{3cm}m{3cm}}

\toprule

Scenario & Optimality gap & Solution time (step 1) & Solution time (step 2) & Total production time & Total processing time &  Total step time \\

\midrule
Scenario I: with delivery time & 0\% & 8.03 sec & 87.24 sec & 8,941 min & 5,963 min & 2,978 min\\
 
Scenario II: without delivery time & 1.26\% & 6.07 sec & 1 hour & 8,627 min & 5,963 min &  2,774 min\\
\bottomrule
\end{tabular}}
\end{table}

\subsubsection{Analysis of personnel shift scheduling} 
\begin{figure}[H]
    \centering
    \includegraphics[width=10cm]{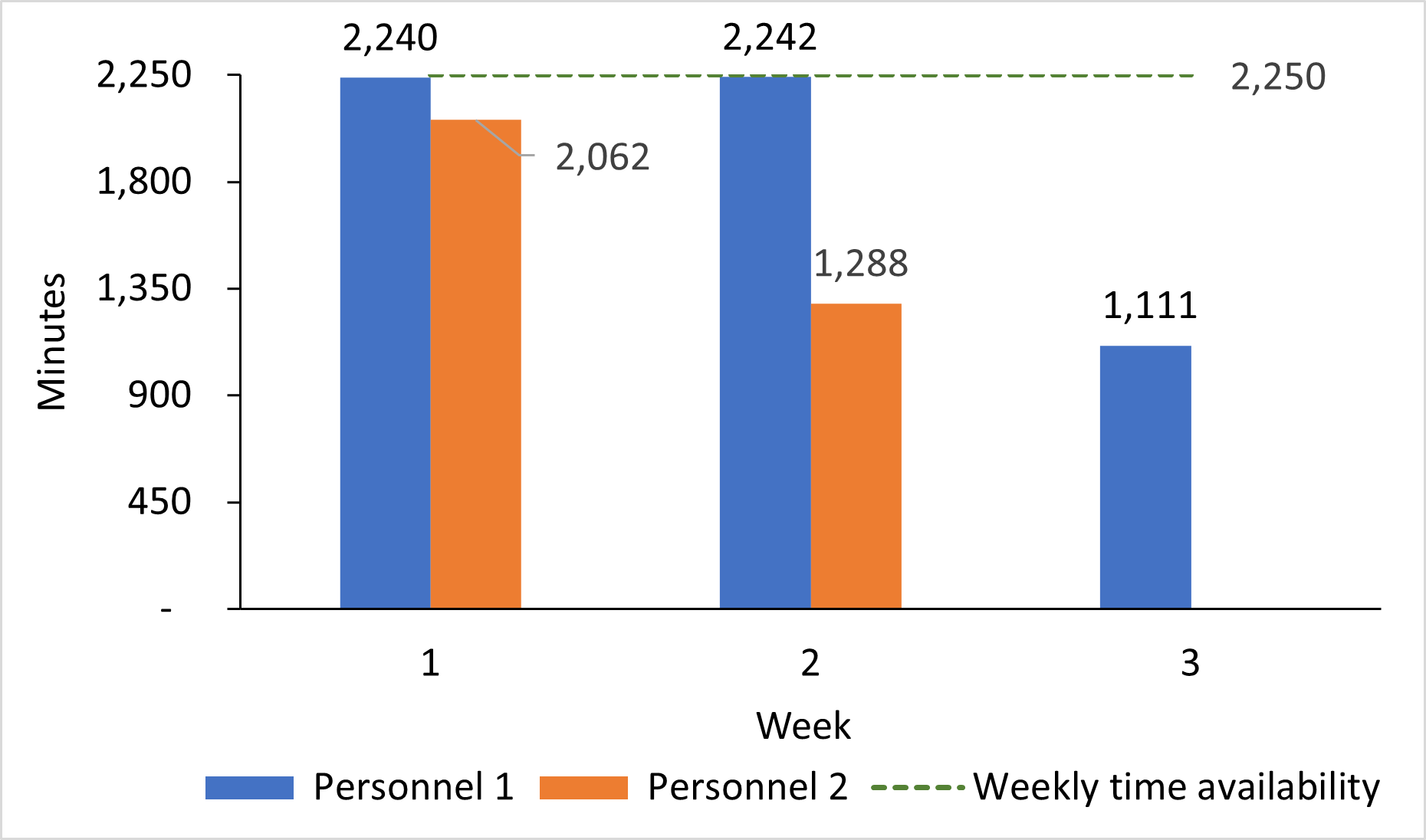}
    \caption{Weekly utilization level of personnel}
    \label{fig:utilization}
\end{figure}

In this section, we investigate the last suggestion proposed above on personnel days off. The primary contribution of our proposed model lies in its consideration of personnel availability alongside production scheduling. This personnel availability is specifically related to the working days of the personnel in our case study. Our aim here is to harness this aspect of the model to explore feasible scenarios for providing personnel with some days off while minimizing the impact on the total production time. In the case study, the planning horizon spans three weeks, with each week comprising five workdays. With two available personnel to operate machines, there is a total of 30 workdays available across the planning horizon. As depicted in \Cref{fig:utilization}, it is evident that personnel remain unutilized for nine full days out of these 30 days. Therefore, they only need to work for 21 days to ensure the production of all customer orders while adhering to their respective deadlines. Although allowing for nine days off-time, the model was run with varying personnel time availability within every week. Infeasible scenarios where one or more orders could not be scheduled are not reported here. Given that the case study delivery times are on a weekly basis, taking any days off within the week would yield the same solution. For this reason, we can adjust personnel availability by modifying the value of the parameter $End_{kt}^P$, which represents the end of the working time of personnel $k$ during time period $t$. This adjustment impacts the constraint outlined in \Cref{eq:t28} by reducing personnel availability within week $t$. Even though the days are reduced from the end of weekly work time for personnel, off days within week $t$ can fluctuate between the first and last workdays of that week.\\

\Cref{tab:personnelhift} presents the potential scenarios. After running the model with various personnel shift schedules, nine feasible scenarios were identified, each with a total production time no greater than 21 days. Scenarios with production times exceeding 21 days were disregarded due to being unjustifiable for the plant to execute, involving unnecessary longer production times, higher production costs, and additional energy consumption. Compared to the base case scenario, where two personnel are available for each of the five days per week and 30 days in total during the planning horizon, in other scenarios, two personnel's availability will be 21 days in total. Their cumulative weekly work days indicate that they should be available for a minimum of eight days, six days, and three days in the first, second, and third week, respectively. However, this minimum weekly availability of personnel depends on the scenario. In Scenario A, the production time remains the same as in the base case. This is because, in this scenario, the idle days of personnel were removed from the total days without changing the generated production schedule. Other scenarios were generated by changing the weekly available days in scenario A and then running the model. As a result, scenarios B to I experience an increase in total production time between 10 minutes and 139 minutes. This increase can be attributed to changes in the production sequence to maintain the production schedule feasible. Consequently, changing the production sequence will affect and increase the setup time. Despite the minor increase in the total production time, personnel utilization improved significantly from 66\% in the base case to well over 95\% in all scenarios. In terms of solution time, it takes longer to solve both steps in all scenarios than in the base case, but it does not exceed 10 minutes in total. The increase in solution time can be justified by the fact that reducing the availability of personnel from 30 to 21 days makes it harder for the model to find a feasible optimal solution. In conclusion, based on the minimum days needed to meet customer demands, we devised the minimum number of days that personnel should cover during every week in different scenarios. These shift schedules led to a considerable improvement in the percentage of personnel utilization, while helping the scheduler to find the best time to allow personnel to take days off with minimal impact on the total production time.\\








\begin{table}[ht]
\caption{Possible scenarios for personnel shift schedules}
\label{tab:personnelhift}
\resizebox{\columnwidth}{!}{
\centering

\begin{tabular}{m{2cm}m{2.75cm}m{2.25cm}m{2.25cm}m{2.25cm}m{3cm}m{3cm}m{3cm}m{3cm}}

\toprule

Scenario & Cumulative weekly work days of personnel & Total work days & Solution time (sec, step 1) & Solution time (sec, step 2) & Total production time (min) & Total available time (min) & Production time increase (min, compared to the base case) & Worker utilization (\%) \\

\midrule
Base case & [10, 10, 10] &	30 &    8.03	& 87.24	 & 8,941 	 & 13,500 	& -	  & 66\% \\
A &         [10, 8, 3] &	21 &	50.2	& 468.47	 & 8,941 	 & 9,450 	& 0	  & 95\% \\
B &         [10, 7, 4] &	21 &    70.31	& 321.48	 & 8,951 	 & 9,450 	& 10  &	95\% \\
C &         [10, 6, 5] &	21 &    25.81	& 127.06	 & 9,042 	 & 9,450 	& 101 &	96\% \\
D &         [9, 9, 3]  &	21 &    20.23	& 250.01	 & 8,956 	 & 9,450 	& 15  & 95\% \\
E &         [9, 8, 4]  &	21 &    329.17	& 171.23	 & 8,966 	 & 9,450 	& 25  & 95\% \\
F &         [9, 7, 5]  &    21 &    130.50	& 325.3	     & 8,985 	 & 9,450 	& 44  &	95\% \\
G &         [8, 9, 4]  &    21 &    116.39	& 179.92	 & 9,061 	 & 9,450 	& 120 & 96\% \\
H &         [8, 10, 3] &    21 &    72.09	& 121.56	 & 9,050 	 & 9,450 	& 109 &	96\% \\
I &         [8, 8, 5]  &    21 &    86.78	& 302.52	 & 9,080 	 & 9,450 	& 139 &	96\% \\

\bottomrule
\end{tabular}}
\end{table}



\section{Conclusions} \label{sec: conclusions}

In this paper, we developed a Mixed Integer Linear Programming (MILP) model for integrated production and personnel shift scheduling in an unrelated parallel machine environment. Unlike previous works that assumed constant resource availability, our model accommodates variations in the work shift hours of each personnel. This is achieved through a discrete multi-period scheduling horizon and continuous time duration for each period, allowing for adjustable personnel availability based on start and end work times within each period. In addition, the model considers common manufacturing settings, such as machine and job sequence-dependent setup times, machine eligibility constraints, job release times, and delivery times. To enhance the solving speed, we introduced a Two-Step Solution Approach (TSSA), providing a warm-start initial solution for the MILP solver. In the first step, TSSA relaxes job allocation constraints to maximize the number of scheduled jobs, followed by minimizing total production time in the second step. The model performance was evaluated on synthetically generated problem instances with up to 120 jobs, eight machines, and seven personnel. The computational results suggest that the number of time periods, machines, and available personnel are the top three influential factors contributing to longer solution times. Furthermore, on average, the findings reveal that the presence of machine eligibility and production time window constraints can increase the total production time by 46\% and 40\%, respectively, compared to the problem instances where these constraints do not exist.

We also applied the model to a scheduling problem encountered in a food processing plant case study, demonstrating its practical effectiveness. The results show that the proposed model can achieve optimal solutions in a short time frame for real-world problem instances. Specifically, the practical utility of the proposed method is that it facilitates calculation of worker utilization by varying shift schedules, e.g., 66\% in base case scenario versus over 95\% in adjusted scenarios. Thus, the proposed model can reduce shift hours while meeting customer orders and delivery times. Given that the model spends significant amount of time finding a feasible solution in the first step of TSSA, future research can focus on designing heuristics to generate warm start feasible solutions, improving computational time and scalability. Furthermore, a metaheuristic solution method can be used to approximate some sub-optimal solutions, and their performance can be compared with our MILP model. Lastly, considering flexibility in resource requirements per machine and more than one additional resource type present other avenues for future research.

\section*{Acknowledgement}
We would like to acknowledge the financial support of NTWIST Inc. and Natural Sciences and Engineering Research Council (NSERC) Canada under the Alliance Grant ALLRP 555220 – 20, and research collaboration of NTWIST Inc. from Canada, Fraunhofer IEM, D\"{u}spohl Gmbh, and Encoway Gmbh from Germany in this research.

\bibliographystyle{tfcad}
\bibliography{interactcadsample}

\begin{thebibliography}{31}
\newcommand{\enquote}[1]{``#1''}
\providecommand{\natexlab}[1]{#1}
\providecommand{\url}[1]{\normalfont{#1}}
\providecommand{\urlprefix}{}

\bibitem[Abbasi, Pishvaee, and Bairamzadeh(2020)]{abbasi2020land}
Abbasi, Mostafa, Mir~Saman Pishvaee, and Samira Bairamzadeh. 2020. ``Land suitability assessment for Paulownia cultivation using combined GIS and Z-number DEA: A case study.'' \emph{Computers and Electronics in Agriculture} 176: 105666.

\bibitem[Afzalirad and Rezaeian(2016)]{afzalirad2016resource}
Afzalirad, Mojtaba, and Javad Rezaeian. 2016. ``Resource-constrained unrelated parallel machine scheduling problem with sequence dependent setup times, precedence constraints and machine eligibility restrictions.'' \emph{Computers \& Industrial Engineering} 98: 40--52.

\bibitem[Ahang et~al.(2024)]{ahang2024intelligent}
Ahang, Maryam, Todd Charter, Oluwaseyi Ogunfowora, Maziyar Khadivi, Mostafa Abbasi, and Homayoun Najjaran. 2024. ``Intelligent Condition Monitoring of Industrial Plants: An Overview of Methodologies and Uncertainty Management Strategies.'' \emph{arXiv preprint arXiv:2401.10266} .

\bibitem[{AIMMS 4.92.7 software}(2023)]{AIMMS}
{AIMMS 4.92.7 software}. 2023. ``{}.'' \url{https://www.aimms.com/}. {Version 4.92.7}.

\bibitem[Al-Harkan and Qamhan(2019)]{al2019optimize}
Al-Harkan, Ibrahim~M, and Ammar~A Qamhan. 2019. ``Optimize unrelated parallel machines scheduling problems with multiple limited additional resources, sequence-dependent setup times and release date constraints.'' \emph{IEEE Access} 7: 171533--171547.

\bibitem[Arnaout, Rabadi, and Musa(2010)]{arnaout2010two}
Arnaout, Jean-Paul, Ghaith Rabadi, and Rami Musa. 2010. ``A two-stage ant colony optimization algorithm to minimize the makespan on unrelated parallel machines with sequence-dependent setup times.'' \emph{Journal of Intelligent Manufacturing} 21: 693--701.

\bibitem[Avalos-Rosales, Angel-Bello, and Alvarez(2015)]{avalos2015efficient}
Avalos-Rosales, Oliver, Francisco Angel-Bello, and Ada Alvarez. 2015. ``Efficient metaheuristic algorithm and re-formulations for the unrelated parallel machine scheduling problem with sequence and machine-dependent setup times.'' \emph{The International Journal of Advanced Manufacturing Technology} 76: 1705--1718.

\bibitem[Avgerinos et~al.(2023)]{avgerinos2023scheduling}
Avgerinos, Ioannis, Ioannis Mourtos, Stavros Vatikiotis, and Georgios Zois. 2023. ``Scheduling unrelated machines with job splitting, setup resources and sequence dependency.'' \emph{International Journal of Production Research} 61 (16): 5502--5524.

\bibitem[Bektur and Sara{\c{c}}(2019)]{bektur2019mathematical}
Bektur, Gulcin, and Tugba Sara{\c{c}}. 2019. ``A mathematical model and heuristic algorithms for an unrelated parallel machine scheduling problem with sequence-dependent setup times, machine eligibility restrictions and a common server.'' \emph{Computers \& Operations Research} 103: 46--63.

\bibitem[Bitar, Dauz{\`e}re-P{\'e}r{\`e}s, and Yugma(2021)]{bitar2021unrelated}
Bitar, Abdoul, St{\'e}phane Dauz{\`e}re-P{\'e}r{\`e}s, and Claude Yugma. 2021. ``Unrelated parallel machine scheduling with new criteria: Complexity and models.'' \emph{Computers \& Operations Research} 132: 105291.

\bibitem[B{\l}a{\.z}ewicz et~al.(2007)]{blazewicz2007handbook}
B{\l}a{\.z}ewicz, Jacek, Klaus~H Ecker, Erwin Pesch, G{\"u}nter Schmidt, and Jan Weglarz. 2007. \emph{Handbook on scheduling: from theory to applications}. Springer Science \& Business Media.

\bibitem[Burdett et~al.(2021)]{burdett2021scheduling}
Burdett, Robert~L, Paul Corry, Colin Eustace, and Simon Smith. 2021. ``Scheduling pre-emptible tasks with flexible resourcing options and auxiliary resource requirements.'' \emph{Computers \& Industrial Engineering} 151: 106939.

\bibitem[Castro, Grossmann, and Novais(2006)]{castro2006two}
Castro, Pedro~M, Ignacio~E Grossmann, and Augusto~Q Novais. 2006. ``Two new continuous-time models for the scheduling of multistage batch plants with sequence dependent changeovers.'' \emph{Industrial \& engineering chemistry research} 45 (18): 6210--6226.

\bibitem[Chen et~al.(2022)]{chen2022unrelated}
Chen, Haichao, Peng Guo, Jesus Jimenez, Zhijie~Sasha Dong, and Wenming Cheng. 2022. ``Unrelated parallel machine photolithography scheduling problem with dual resource constraints.'' \emph{IEEE Transactions on Semiconductor Manufacturing} 36 (1): 100--112.

\bibitem[Costa, Cappadonna, and Fichera(2013)]{costa2013hybrid}
Costa, A, FA~Cappadonna, and S~Fichera. 2013. ``A hybrid genetic algorithm for job sequencing and worker allocation in parallel unrelated machines with sequence-dependent setup times.'' \emph{The International Journal of Advanced Manufacturing Technology} 69: 2799--2817.

\bibitem[Daniels, Hoopes, and Mazzola(1996)]{daniels1996scheduling}
Daniels, Richard~L, Barbara~J Hoopes, and Joseph~B Mazzola. 1996. ``Scheduling parallel manufacturing cells with resource flexibility.'' \emph{Management science} 42 (9): 1260--1276.

\bibitem[Daniels, Hoopes, and Mazzola(1997)]{daniels1997analysis}
Daniels, Richard~L, Barbara~J Hoopes, and Joseph~B Mazzola. 1997. ``An analysis of heuristics for the parallel-machine flexible-resource scheduling problem.'' \emph{Annals of Operations Research} 70 (0): 439--472.

\bibitem[Edis, Oguz, and Ozkarahan(2013)]{edis2013parallel}
Edis, Emrah~B, Ceyda Oguz, and Irem Ozkarahan. 2013. ``Parallel machine scheduling with additional resources: Notation, classification, models and solution methods.'' \emph{European Journal of Operational Research} 230 (3): 449--463.

\bibitem[Fanjul-Peyro(2020)]{fanjul2020models}
Fanjul-Peyro, Luis. 2020. ``Models and an exact method for the unrelated parallel machine scheduling problem with setups and resources.'' \emph{Expert Systems with Applications: X} 5: 100022.

\bibitem[Fanjul-Peyro, Ruiz, and Perea(2019)]{fanjul2019reformulations}
Fanjul-Peyro, Luis, Rub{\'e}n Ruiz, and Federico Perea. 2019. ``Reformulations and an exact algorithm for unrelated parallel machine scheduling problems with setup times.'' \emph{Computers \& Operations Research} 101: 173--182.

\bibitem[{Gurobi 10.0.2 software}(2023)]{Gurobi}
{Gurobi 10.0.2 software}. 2023. ``{}.'' \url{https://www.gurobi.com/}. {Version 10.0.2}.

\bibitem[Khadivi et~al.(2023)]{khadivi2023deep}
Khadivi, Maziyar, Todd Charter, Marjan Yaghoubi, Masoud Jalayer, Maryam Ahang, Ardeshir Shojaeinasab, and Homayoun Najjaran. 2023. ``Deep reinforcement learning for machine scheduling: Methodology, the state-of-the-art, and future directions.'' \emph{arXiv preprint arXiv:2310.03195} .

\bibitem[Liang et~al.(2022)]{liang2022lenovo}
Liang, Yi, Zan Sun, Tianheng Song, Qiang Chou, Wei Fan, Jianping Fan, Yong Rui, et~al. 2022. ``Lenovo Schedules Laptop Manufacturing Using Deep Reinforcement Learning.'' \emph{INFORMS Journal on Applied Analytics} 52 (1): 56--68.

\bibitem[Ogunfowora and Najjaran(2023)]{ogunfowora2023reinforcement}
Ogunfowora, Oluwaseyi, and Homayoun Najjaran. 2023. ``Reinforcement and deep reinforcement learning-based solutions for machine maintenance planning, scheduling policies, and optimization.'' \emph{Journal of Manufacturing Systems} 70: 244--263.

\bibitem[Shojaeinasab et~al.(2022)]{shojaeinasab2022intelligent}
Shojaeinasab, Ardeshir, Todd Charter, Masoud Jalayer, Maziyar Khadivi, Oluwaseyi Ogunfowora, Nirav Raiyani, Marjan Yaghoubi, and Homayoun Najjaran. 2022. ``Intelligent manufacturing execution systems: A systematic review.'' \emph{Journal of Manufacturing Systems} 62: 503--522.

\bibitem[S{\l}owi{\'n}ski(1980)]{slowinski1980two}
S{\l}owi{\'n}ski, Roman. 1980. ``Two approaches to problems of resource allocation among project activities—a comparative study.'' \emph{Journal of the Operational Research Society} 31 (8): 711--723.

\bibitem[{Tableau Desktop 2023.1.2 Professional Edition}(2023)]{Tableau}
{Tableau Desktop 2023.1.2 Professional Edition}. 2023. ``{}.'' \url{https://www.tableau.com/products/desktop/download}. {Version 2023.1.2}.

\bibitem[Tahmassebi(1999)]{tahmassebi1999vehicle}
Tahmassebi, T. 1999. ``Vehicle routing problem (VRP) formulation for continuous-time packing hall design/operations.'' \emph{Computers \& Chemical Engineering} 23: S1011--S1014.

\bibitem[Tran, Araujo, and Beck(2016)]{tran2016decomposition}
Tran, Tony~T, Arthur Araujo, and J~Christopher Beck. 2016. ``Decomposition methods for the parallel machine scheduling problem with setups.'' \emph{INFORMS Journal on Computing} 28 (1): 83--95.

\bibitem[Yepes-Borrero et~al.(2020)]{yepes2020grasp}
Yepes-Borrero, Juan~C, Fulgencia Villa, Federico Perea, and Juan~Pablo Caballero-Villalobos. 2020. ``GRASP algorithm for the unrelated parallel machine scheduling problem with setup times and additional resources.'' \emph{Expert Systems with Applications} 141: 112959.

\bibitem[Yunusoglu and Topaloglu~Yildiz(2022)]{yunusoglu2022constraint}
Yunusoglu, Pinar, and Seyda Topaloglu~Yildiz. 2022. ``Constraint programming approach for multi-resource-constrained unrelated parallel machine scheduling problem with sequence-dependent setup times.'' \emph{International Journal of Production Research} 60 (7): 2212--2229.

\end{thebibliography}

\appendix

\end{document}